\documentclass{article}
\usepackage{float}

\PassOptionsToPackage{numbers,compress}{natbib}
\usepackage[preprint]{neurips_2026}

\usepackage[utf8]{inputenc} 
\usepackage[T1]{fontenc}    
\usepackage{hyperref}       
\usepackage{url}            
\usepackage{booktabs}       
\usepackage{amsfonts}       
\usepackage{nicefrac}       
\usepackage{microtype}      
\usepackage{xcolor}         
\usepackage{amsmath}
\usepackage{graphicx}
\usepackage{algpseudocode}
\usepackage{float}
\usepackage{multirow}
\usepackage{microtype}
\usepackage{subcaption}
\usepackage{booktabs}
\usepackage{amssymb}
\usepackage{mathtools}
\usepackage{amsthm}
\usepackage{algorithm}
\usepackage{algorithmicx}
\usepackage{algpseudocode}
\usepackage{fontawesome5}
\usepackage[table]{xcolor}
  \usepackage{booktabs, pifont}
  \newcommand{\cmark}{\ding{51}}
  \newcommand{\xmark}{\ding{55}}
\usepackage[capitalize,noabbrev]{cleveref}
\newcommand{\doraem}{\raisebox{-0.2ex}{\includegraphics[height=1em]{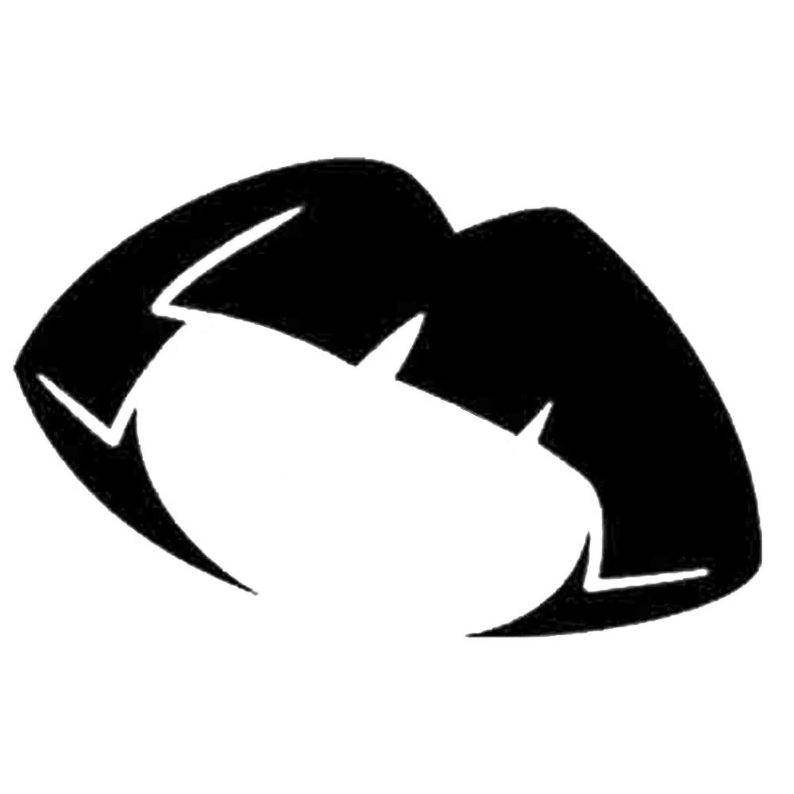}}}
\newcommand{\mickeyem}{\raisebox{-0.2ex}{\includegraphics[height=1em]{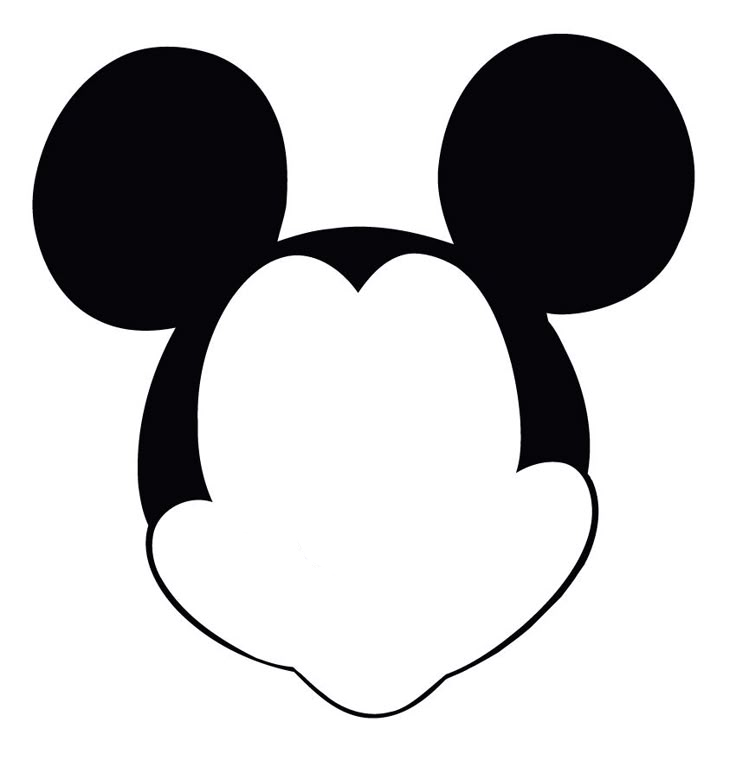}}}
\newcommand{\legendsq}[1]{{\setlength{\fboxsep}{0pt}\fcolorbox{black!30}{#1}{\rule{0pt}{1.4ex}\rule{1.4ex}{0pt}}}}
\newcommand{\bluearrow}{\raisebox{-0.2ex}{\includegraphics[height=1em]{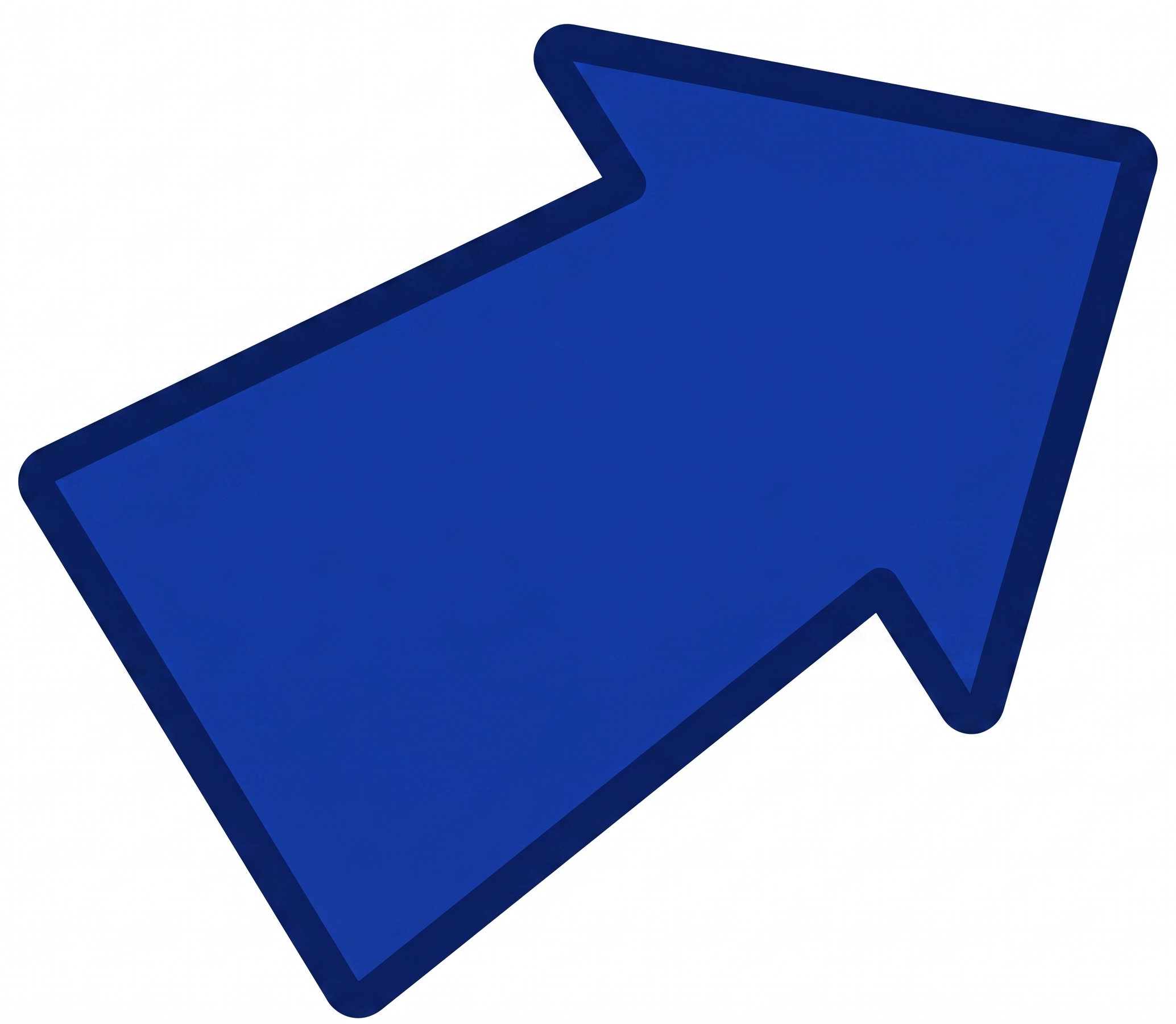}}}

\usepackage{listings}
\usepackage{tcolorbox}
\tcbuselibrary{listings,skins,breakable}
\definecolor{promptbg}{RGB}{248,249,252}
\definecolor{promptframe}{RGB}{70,90,140}
\definecolor{promptkey}{RGB}{170,40,90}
\definecolor{promptstr}{RGB}{30,110,60}
\definecolor{promptcom}{RGB}{120,120,120}
\definecolor{promptplaceholder}{RGB}{200,90,30}

\newcommand{\sosvqafigure}{%
\begin{figure*}[t]
\centering
\includegraphics[width=0.85\textwidth]{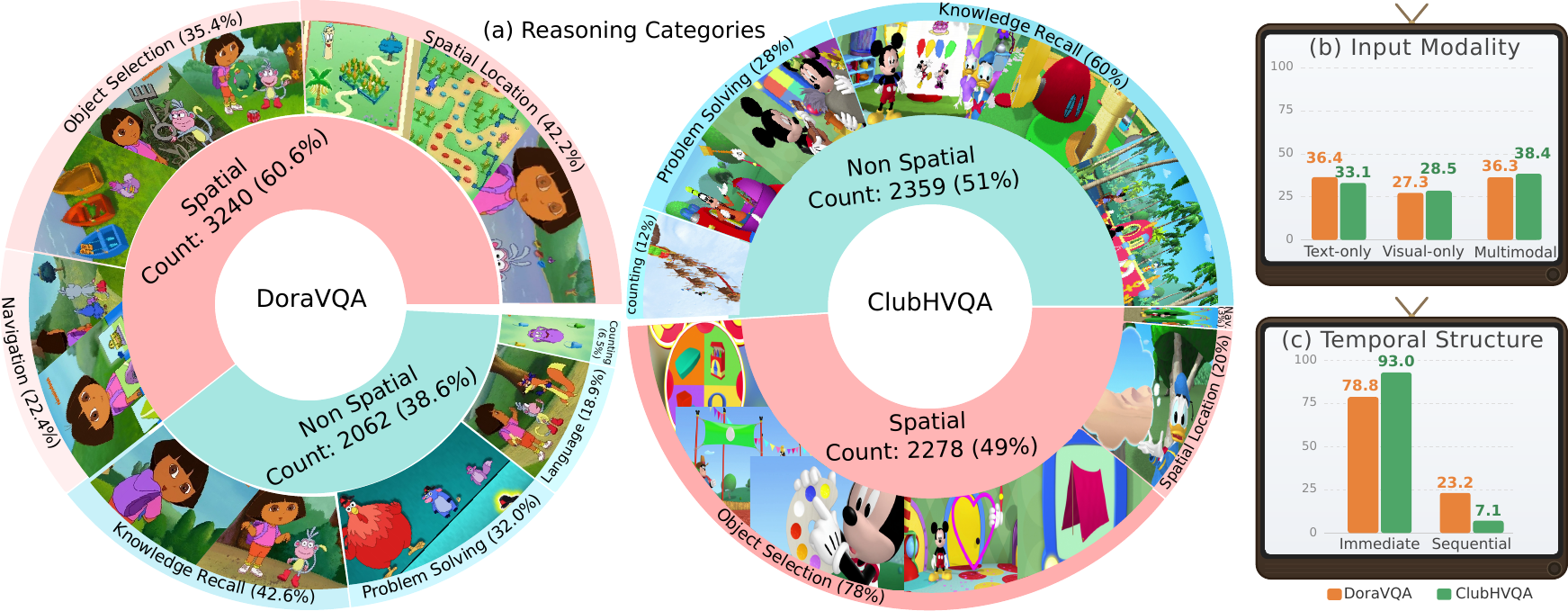}
\caption{\textbf{SoSVQA} (Structure over Scale Visual Question Answering) dataset composition across two subsets: DoraVQA and ClubHVQA. \textbf{(a) Reasoning categories:} DoraVQA is dominated by spatial-reasoning tasks, whereas ClubHVQA emphasizes object-selection and knowledge-recall tasks. \textbf{(b) Input modality:} Both datasets exhibit similar input modality distributions. \textbf{(c) Temporal structure:} ClubHVQA predominantly requires immediate reasoning, whereas DoraVQA includes a substantially larger portion of sequential-reasoning questions.
} 
\label{fig:sosvqa-overview}
\vspace{-0.8cm}
\end{figure*}
}

\newcommand{\doravqapipeline}{%
\begin{figure*}[t]
\centering
\includegraphics[width=0.95\textwidth]{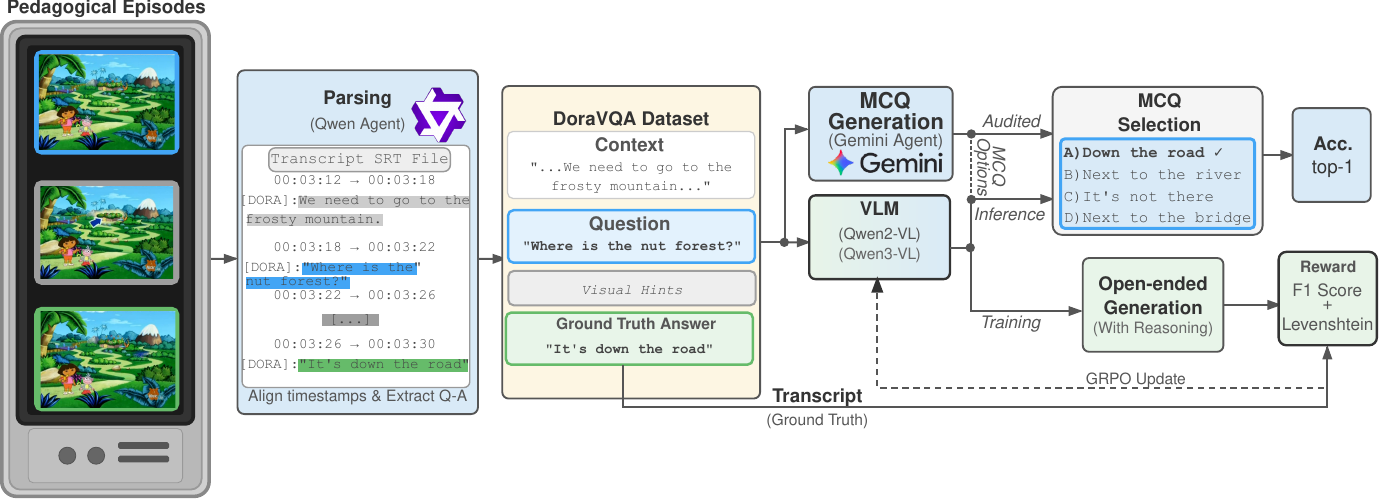}
\caption{\textbf{SoSVQA pipeline overview.} We extract question-answer pairs from
pedagogical episodes by parsing SubRip Text (SRT) transcript files with a Qwen agent, aligning timestamps to identify the show's pedagogical \textit{context-question-pause-answer} structure. Each question is paired with its surrounding context window, the ground truth answer, and multiple-choice distractors (MCQs) that follows. During training, we fine-tune Qwen models using GRPO on open-ended generation; during inference, the MCQ options are provided to our fine-tuned models, creating a deliberate train-test format mismatch (open-ended $\rightarrow$ MCQ) that evaluates transferable reasoning.}
\label{fig:doravqa-pipeline}
\vspace{-0.5cm}
\end{figure*}
}

\newcommand{\quality}{%
\begin{figure*}[t]
\centering
\includegraphics[width=1.0\textwidth]{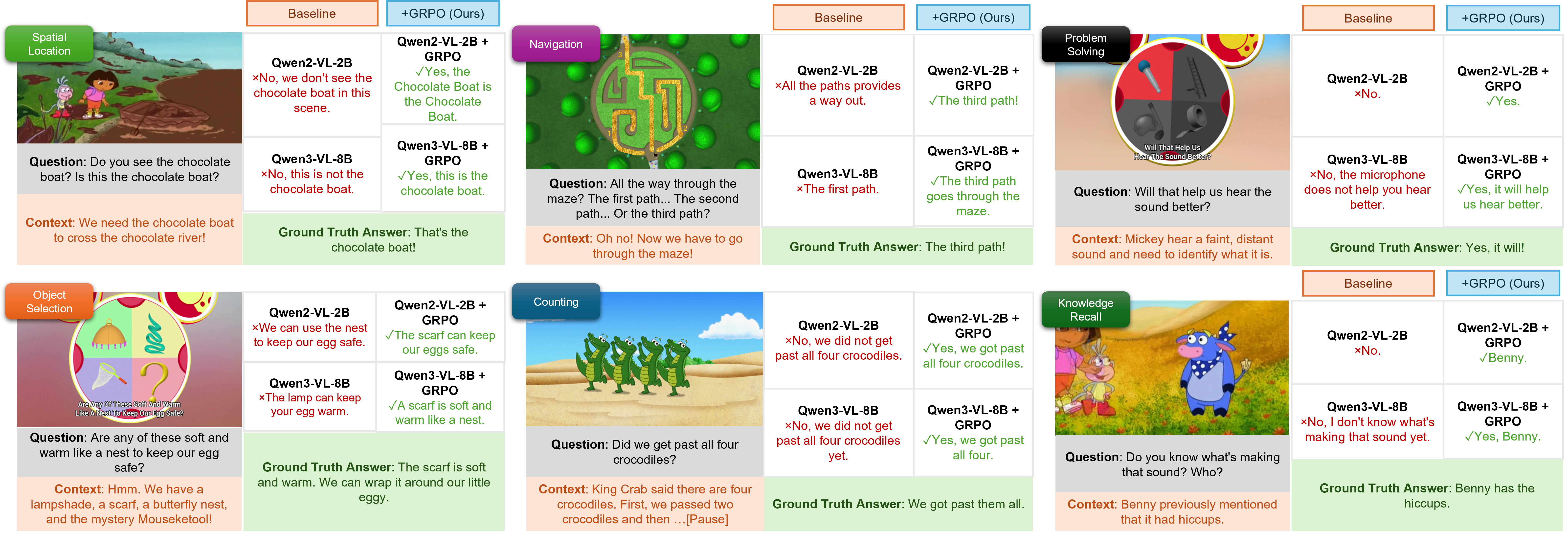}
\caption{\textbf{Qualitative comparison on some samples of challenging spatial reasoning tasks.} Our GRPO‑finetuned models (blue) outperform all baselines (orange) across every reasoning category. While baseline models frequently fail or hallucinate, our models consistently produce accurate and concise answers. See Appendix~\ref{sec:additional_qualitative} and supplementary material for more examples.
}
\label{fig:doravqa-qualitative}
\vspace{-0.5cm}
\end{figure*}
}

\newcommand{\additionalquality}{%
\begin{figure*}[htbp]
\centering
\includegraphics[width=\textwidth]{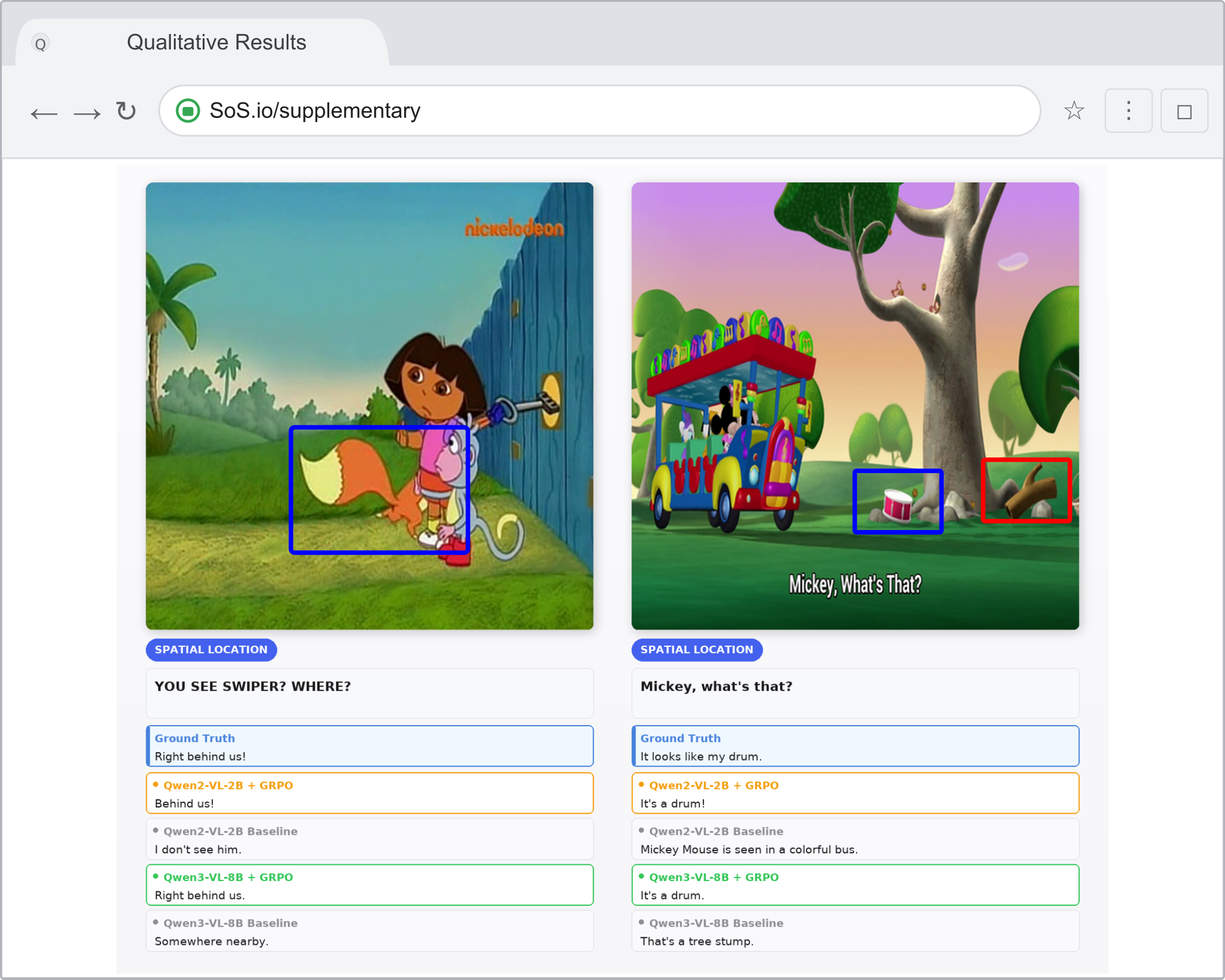}
\caption{\textbf{Additional challenging examples from SoSVQA.} Left: Spatial location task where Swiper the fox is partially occluded behind a blue wall, requiring the model to recognize objects from incomplete visual information (blue box). Right: A deixis resolution task where the question ``Mickey, what's that?'' requires understanding \textit{what} Mickey is referring to within the episodic narrative context. Baseline models either describe the dominant visual element (the colorful bus) or latch onto an unrelated object---the tree stump (red box)---revealing that they answer ``what is visually prominent?'' rather than ``what is Mickey referring to?''. Our GRPO-finetuned models correctly resolve the reference to the drum (blue box) by grounding the question in the episode's narrative context. \textbf{Additional interactive qualitative results.} We provide an interactive HTML page in the supplementary materials with further visual examples across all task types. Please open \texttt{supplementary/index.html} in a web browser to explore the full set of qualitative results beyond the examples shown in this paper.}
\label{fig:additional-qualitative}
\end{figure*}
}

\newcommand{\pausehintquality}{%
\begin{figure*}[t]
\centering
\includegraphics[width=0.95\textwidth]{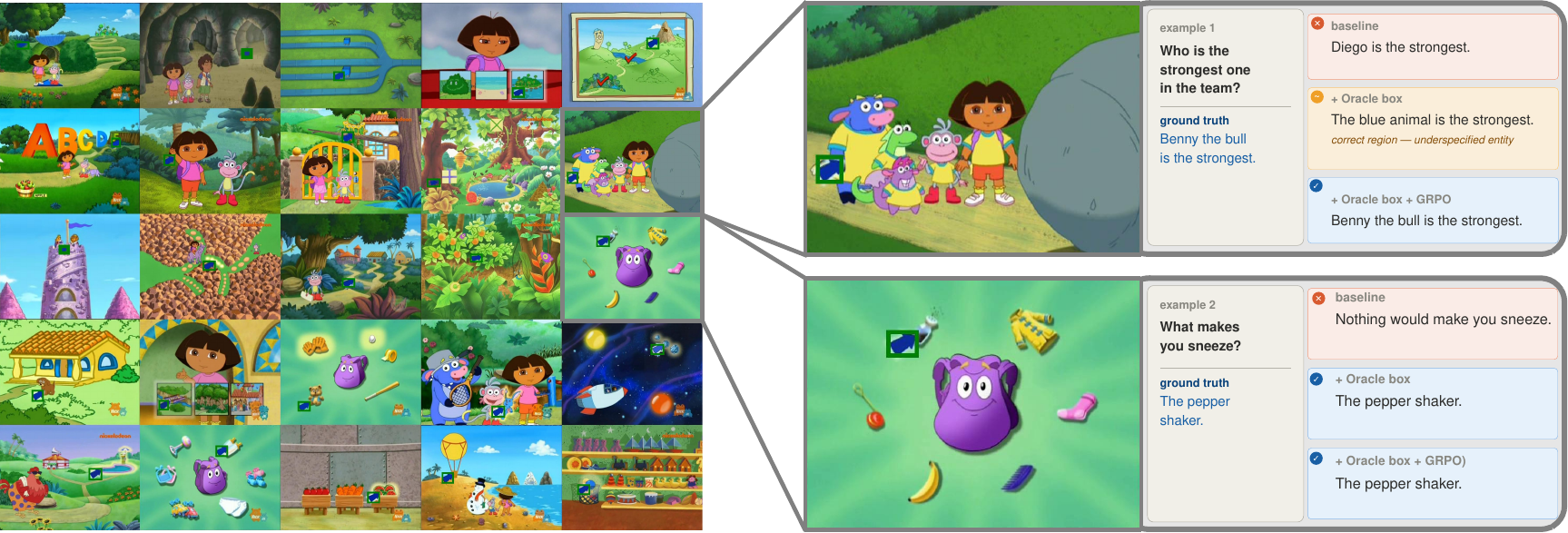}
\caption{\textbf{Effect of pause-based visual hints on spatial grounding.}
We evaluate three conditions on two examples drawn from \textit{Dora the Explorer}'s
pedagogical pause segments, where a blue arrow highlights the visually salient object
relevant to the question.
\textbf{(No hint)} receives only the raw frame and question with no additional guidance.
\textbf{(+ Oracle box)} additionally supplies the ground-truth bounding box of the
arrow's target region as part of the prompt.
\textbf{(+ Oracle box + GRPO)} combines the same oracle box with our GRPO-finetuned model.
In \textit{Example~1}, the baseline hallucinates an incorrect character (Diego); the
oracle box redirects attention to the correct spatial region but the untuned model
describes the referent only at the level of a perceptual property
(``the blue animal''), failing to resolve it to a named entity. Only GRPO fine-tuning
enables the model to bridge from spatial location to the correct named answer
(Benny the bull).
In \textit{Example~2}, both conditions succeed, confirming that an unambiguous spatial
hint is sufficient for object-level queries that require no entity-level knowledge.
Together, these results demonstrate that the show's pause arrow functions as an
\emph{implicit reasoning trace}: it constrains the visual search space, but
generalizing from spatial localization to grounded entity recognition requires the
structured supervision provided by GRPO training on pedagogical content.}
\label{fig:pause-hint-qualitative}
\end{figure*}
}

\newcommand{\mcqauditfigure}{%
\begin{figure}[t]
\centering
\begin{minipage}[c]{0.34\linewidth}
  \centering
  \includegraphics[width=\linewidth]{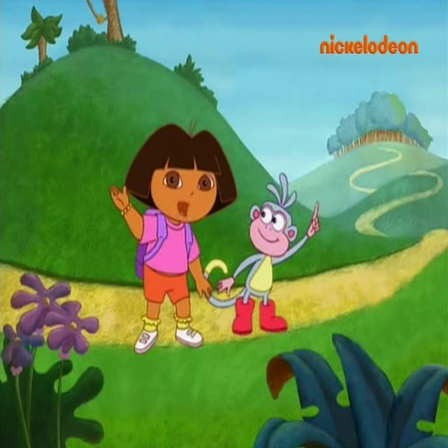}\\[3pt]
  {\footnotesize\itshape Pause segment: Boots is pointing to the quiet forest on the right.}
\end{minipage}%
\hspace{0.015\linewidth}%
\begin{minipage}[c]{0.63\linewidth}
\small
\textbf{Question:}\ \textit{Now, where's the quiet forest?\ Do you see the quiet forest?}\\[2pt]
\textbf{Gold answer:}\ \textit{Yes-- there it is} $\;\rightarrow\;$ \textit{There it is!}

\vspace{5pt}
\setlength{\fboxsep}{5pt}
\fcolorbox{red!45}{red!4}{%
\begin{minipage}{\dimexpr\linewidth-2\fboxsep-2\fboxrule}
\textbf{\textcolor{red!65!black}{Before audit\,:\,shortcut leak}}
\begin{itemize}\setlength{\itemsep}{0pt}\setlength{\topsep}{3pt}\setlength{\parsep}{0pt}\setlength{\leftmargini}{14pt}
  \item Little bird?
  \item Yellow Valley?
  \item \textbf{There it is!}~\cmark
  \item Big mountain?
\end{itemize}
\vspace{2pt}
{\footnotesize\itshape The deictic gold answer is the only non-entity option---selectable by surface form alone.}
\end{minipage}}

\vspace{5pt}
\fcolorbox{green!55!black}{green!4}{%
\begin{minipage}{\dimexpr\linewidth-2\fboxsep-2\fboxrule}
\textbf{\textcolor{green!45!black}{After audit\,:\,visually grounded}}
\begin{itemize}\setlength{\itemsep}{0pt}\setlength{\topsep}{3pt}\setlength{\parsep}{0pt}\setlength{\leftmargini}{14pt}
  \item Yes, it's on the left.
  \item No, I can't see it.
  \item \textbf{Yes, it's on the right.}~\cmark
  \item I see flowers.
\end{itemize}
\vspace{2pt}
{\footnotesize\itshape All four options share a parallel yes/no--directional surface form, requiring the model to ground the answer in the on-screen arrow.}
\end{minipage}}
\end{minipage}
\caption{Manual audit of deictic gold answers in SoSVQA. The original choice
set leaks the answer through surface form; ``There it is!'' is the only
non-named-entity option and can be selected without attending to the visual
scene. The audited choice set forces visual grounding by making all four
options parallel directional statements. We applied this audit to all 106
deictic items in DoraVQA and 31 in ClubHVQA.}
\label{fig:mcq-audit}
\end{figure}}
\newcommand{\doravqamainresults}{
\begin{table*}[t]
\centering
\caption{Performance comparison on \doraem~DoraVQA and \mickeyem~ClubHVQA test sets across reasoning categories. Our \textbf{+ GRPO} models are fine-tuned jointly on the union of DoraVQA and ClubHVQA training splits.
\legendsq{gray!15}~Gray rows indicate proprietary models;
\underline{underlined} scores indicate results surpassed by our best model
(Qwen3-VL-8B + GRPO).
\textcolor{green!70!black}{Green} indicates improvement over baseline,
\textcolor{red!70!black}{red} indicates degradation. All results are reported using top-1 accuracy.} 
\resizebox{\textwidth}{!}{%
\begin{tabular}{l cc cc cc cc cc}
\toprule
& \multicolumn{2}{c}{\textbf{Overall}}
& \multicolumn{2}{c}{\textbf{Spatial}}
& \multicolumn{2}{c}{\textbf{Obj. Selection \& Counting}}
& \multicolumn{2}{c}{\textbf{Navigation}}
& \multicolumn{2}{c}{\textbf{Knowledge}} \\
\cmidrule(lr){2-3}\cmidrule(lr){4-5}\cmidrule(lr){6-7}\cmidrule(lr){8-9}\cmidrule(lr){10-11}
\textbf{Model}
  & \textbf{\doraem} & \textbf{\mickeyem}
  & \textbf{\doraem} & \textbf{\mickeyem}
  & \textbf{\doraem} & \textbf{\mickeyem}
  & \textbf{\doraem} & \textbf{\mickeyem}
  & \textbf{\doraem} & \textbf{\mickeyem} \\
\midrule
\rowcolor{gray!15}
Gemini-3.0-Flash              & 76.10 & 69.16 & \underline{67.83} & 62.18 & 87.10 & 73.10 & 65.38 & 65.91 & 88.52 & 68.57 \\
\rowcolor{gray!15}
Gemini-3.0-Flash (low thinking) & 75.84 & 63.96 & \underline{66.81} & \underline{54.98} & 88.17 & 68.52 & 69.23 & 61.36 & 86.89 & 60.83 \\
\rowcolor{gray!15}
GPT-4V                        & 67.79 & 68.20 & \underline{60.34} & 64.10 & 75.27 & 69.90 & 69.23 & 65.91 & 68.85 & 66.15 \\
\rowcolor{gray!15}
Gemini-2.5-Pro                & 64.41 & 62.69 & \underline{62.62} & \underline{52.92} & 72.83 & 70.53 & 60.78 & 65.91 & 71.93 & 60.00 \\
\midrule
LLaVA-Video-7B                & \underline{55.41} & \underline{46.46} & \underline{54.02} & \underline{45.10} & \underline{62.22} & \underline{54.52} & 66.00 & \underline{47.62} & 67.31 & \underline{48.66} \\
Video-LLaVA-7B                & \underline{37.82} & \underline{39.19} & \underline{47.49} & \underline{41.48} & \underline{60.66} & \underline{42.78} & \underline{46.34} & \underline{56.76} & \underline{32.43} & \underline{40.68} \\
InternVideo2.5-8B             & \underline{57.68} & \underline{50.10} & \underline{46.55} & \underline{49.52} & \underline{63.04} & \underline{55.87} & 59.62 & \underline{36.36} & 68.85 & \underline{53.14} \\
\midrule
Qwen2-VL-2B (baseline)        & 40.29 & 31.43 & 45.36 & 33.99 & 37.61 & 36.10 & 39.03 & 23.26 & 39.95 & 32.87 \\
\textbf{Qwen2-VL-2B + GRPO}
& \textbf{48.32} \textcolor{green!70!black}{\small(+8.03)} & \textbf{36.68} \textcolor{green!70!black}{\small(+5.25)}
& \textbf{58.86} \textcolor{green!70!black}{\small(+13.50)} & \textbf{45.21} \textcolor{green!70!black}{\small(+11.22)}
& \textbf{44.71} \textcolor{green!70!black}{\small(+7.10)} & \textbf{42.38} \textcolor{green!70!black}{\small(+6.28)}
& \textbf{44.30} \textcolor{green!70!black}{\small(+5.27)} & \textbf{46.51} \textcolor{green!70!black}{\small(+23.25)}
& \textbf{45.97} \textcolor{green!70!black}{\small(+6.02)} & \textbf{38.07} \textcolor{green!70!black}{\small(+5.20)} \\
\midrule
Qwen2-VL-7B (baseline)        & 52.21 & 45.09 & 54.14 & 44.55 & 51.45 & 49.51 & 41.97 & 40.91 & 48.77 & 48.05 \\
\textbf{Qwen2-VL-7B + GRPO}
& \textbf{57.68} \textcolor{green!70!black}{\small(+5.47)} & \textbf{50.10} \textcolor{green!70!black}{\small(+5.01)}
& \textbf{67.25} \textcolor{green!70!black}{\small(+13.11)} & \textbf{50.97} \textcolor{green!70!black}{\small(+6.42)}
& \textbf{64.37} \textcolor{green!70!black}{\small(+12.92)} & \textbf{58.98} \textcolor{green!70!black}{\small(+9.47)}
& \textbf{47.80} \textcolor{green!70!black}{\small(+5.83)} & \textbf{50.00} \textcolor{green!70!black}{\small(+9.09)}
& \textbf{67.25} \textcolor{green!70!black}{\small(+18.48)} & \textbf{54.49} \textcolor{green!70!black}{\small(+6.44)} \\
\midrule
Qwen3-VL-8B (baseline)        & 55.24 & 50.67 & 57.17 & 49.20 & 53.58 & 58.62 & \textbf{48.96} & 45.45 & 50.32 & 50.50 \\
\textbf{Qwen3-VL-8B + GRPO}
& \textbf{62.98} \textcolor{green!70!black}{\small(+7.74)} & \textbf{56.81} \textcolor{green!70!black}{\small(+1.57)}
& \textbf{69.65} \textcolor{green!70!black}{\small(+12.48)} & \textbf{59.16} \textcolor{green!70!black}{\small(+1.99)}
& \textbf{64.59} \textcolor{green!70!black}{\small(+11.01)} & \textbf{64.87} \textcolor{green!70!black}{\small(+6.25)}
& 48.33 \textcolor{red!70!black}{\small(-0.63)} & \textbf{60.74} \textcolor{green!70!black}{\small(+15.29)}
& \textbf{55.12} \textcolor{green!70!black}{\small(+4.80)} & \textbf{57.48} \textcolor{green!70!black}{\small(+6.98)} \\
\bottomrule
\end{tabular}}
\label{tab:doravqa_main}
\end{table*}

}

\newcommand{\crossbenchmarkresults}{
\begin{table*}[t]
\vspace{-0.1cm}
\centering
\small
\caption{Cross-benchmark transfer evaluation highlighting \textbf{structure vs.\ scale}.
Despite fine-tuning on only 10K QA pairs (78 hours), our models achieve competitive or
state-of-the-art results against models trained on orders of magnitude more data.
\legendsq{gray!15}~proprietary models;
\legendsq{yellow!10}~our state-of-the-art result;
\legendsq{blue!8}~our minimal fine-tuning addition.
\textcolor{green!70!black}{Green} indicates improvement over baseline.
$^\dagger$ indicates zero-shot evaluation; * indicates training set seen during
pre-training. Our GRPO models share the same pre-training corpus as their baselines but additionally fine-tune jointly on the 10K pedagogically-structured QA pairs from SoSVQA (78 hours of educational video).}
\label{tab:cross_benchmark}
\resizebox{\textwidth}{!}{%
\begin{tabular}{l c c c c c c c}
\toprule
\textbf{Model} & \textbf{Params} & \textbf{Scale} & \textbf{V-MME} & \textbf{NExT-QA}
  & \textbf{MVBench} & \textbf{MotionBench} & \textbf{TVBench} \\
\midrule
\rowcolor{gray!15}
GPT-4V$^\dagger$          & $\sim$1.8T & $\sim$10T tokens & 59.9  & 68.2  & 43.50   & 58.82     &34.00   \\
\rowcolor{gray!15}
Gemini-2.5-Pro$^\dagger$  & --         & --               & 85.2  & 74.6  & 45.50  & 66.30   & 46.50   \\
\rowcolor{gray!15}
Gemini-3.0-Flash$^\dagger$& --         & --               & \textbf{86.9} & 80.4 & 65.50 & 70.40 & 38.25 \\
\midrule
InternVideo2.5-8B  & 8B & 16M clips    & 65.4  & 71.5*  & 75.7   & 45.56   & 48.80   \\
LLaVA-Video-7B     & 7B & 1.3M videos  & 63.3  & \textbf{83.2}* & 69.7 & 48.29 & 42.30 \\
Video-LLaVA-7B     & 7B & 760K videos  & 45.3  & 52.1*  & 40.95   & 35.72   & 36.30   \\
\midrule
Qwen2-VL-2B (baseline)        & 2B & 1.2T tokens          & 50.10 & 52.60 & 58.99 & 47.0 & 39.85   \\
Qwen2-VL-2B + GRPO$^\dagger$  & 2B & \cellcolor{blue!8}1.2T + \textbf{10K QA} & \textbf{60.74} \textcolor{green!70!black}{\small(+10.64)} & \textbf{72.11} \textcolor{green!70!black}{\small(+19.51)} & \textbf{59.94} \textcolor{green!70!black}{\small(+0.95)} & 51.92 \textcolor{green!70!black}{\small(+4.92)} & 40.05 \textcolor{green!70!black}{\small(+0.80)} \\
\midrule
Qwen2-VL-7B (baseline)        & 7B & 1.2T tokens          & 67.50  & 67.0  & \textbf{65.49}  &  52.0  & \textbf{40.75}   \\
Qwen2-VL-7B + GRPO$^\dagger$  & 7B & \cellcolor{blue!8}1.2T + \textbf{10K QA} & \textbf{69.25} \textcolor{green!70!black}{\small(+1.75)} & \textbf{81.80} \textcolor{green!70!black}{\small(+12.46)} & 64.33 \textcolor{red!70!black}{\small(-1.16)} & \textbf{54.01} \textcolor{green!70!black}{\small(+2.01)} & 40.60 \textcolor{red!70!black}{\small(-0.15)}\\
\midrule
Qwen3-VL-8B (baseline)        & 8B & 1.0T tokens          & 71.40  & 62.1  & \textbf{68.70} & \textbf{61.12} & 46.53   \\
\rowcolor{yellow!10}
Qwen3-VL-8B + GRPO$^\dagger$  & 8B & \cellcolor{blue!8}1.0T + \textbf{10K QA} & \textbf{80.00} \textcolor{green!70!black}{\small(+8.60)} & \textbf{81.80}$^\dagger$ \textcolor{green!70!black}{\small(+19.70)} & 67.59 \textcolor{red!70!black}{\small(-1.11)}   & 61.03 \textcolor{red!70!black}{\small(-0.09)} & 47.48 \textcolor{green!70!black}{\small(+0.95)} \\
\bottomrule
\end{tabular}}%
\vspace{-0.1cm}
\end{table*}
}

\newcommand{\ablationtrainingmethod}{%
\begin{table}[t]
\vspace{-0.1cm}
\centering
\footnotesize
\setlength{\tabcolsep}{4pt}
\renewcommand{\arraystretch}{0.9}
\setlength{\aboverulesep}{1pt}
\setlength{\belowrulesep}{1pt}
\caption{Training-data ablations.
\textbf{(a)} Cross-show transfer with Qwen2-VL-2B: a model trained on one
SoSVQA split is evaluated on both, plus three external benchmarks.
\textbf{(b)} Generic (NExT-OE) training, 10K QA, across three backbones; the
gain over the SoSVQA-joint training reported in
Tables~\ref{tab:doravqa_main} and~\ref{tab:cross_benchmark} isolates the
contribution of pedagogical structure.
\legendsq{gray!15}~shaded cells denote in-domain evaluation.
$^*$~indicates eval set seen during fine-tuning.}
\label{tab:ablation_method}
\vspace{4pt}
\resizebox{\textwidth}{!}{%
\begin{tabular}{@{}ll|cc|cc@{\hskip 12pt}|@{\hskip 12pt}l|cc|cc@{}}
\toprule
\multicolumn{6}{c}{\textbf{(a) Cross-show transfer (Qwen2-VL-2B)}}
& \multicolumn{5}{c}{\textbf{(b) Generic training (NExT-OE, 10K QA)}} \\
\cmidrule(r){1-6} \cmidrule(l){7-11}
\multirow{2}{*}{\textbf{Train}} & \multirow{2}{*}{\textbf{Method}}
& \multicolumn{2}{c|}{\textbf{In-Domain}}
& \multicolumn{2}{c}{\textbf{Cross-Benchmark}}
& \multirow{2}{*}{\textbf{Model}}
& \multicolumn{2}{c|}{\textbf{In-Domain}}
& \multicolumn{2}{c}{\textbf{Cross-Benchmark}} \\
\cmidrule(lr){3-4}\cmidrule(lr){5-6}\cmidrule(lr){8-9}\cmidrule(lr){10-11}
& & \doraem & \mickeyem & V-MME & NExT
& & \doraem & \mickeyem & V-MME & NExT \\
\midrule
--        & Baseline      & 41.36 & 50.35 & 50.10 & 52.60
& Qwen2-VL-2B & 40.11 & 30.67 & 51.55 & \cellcolor{gray!15}72.00$^*$ \\
\doraem   & SFT           & 54.30 & 30.52 & 53.33 & 66.63
& Qwen2-VL-7B & 53.55 & 44.30 & 65.33 & \cellcolor{gray!15}82.33$^*$ \\
\doraem   & \textbf{GRPO} & \cellcolor{gray!15}\textbf{55.11} & \textbf{54.45} & \textbf{55.40} & \textbf{72.86}
& Qwen3-VL-8B & \textbf{56.20} & \textbf{50.75} & \textbf{74.44} & \cellcolor{gray!15}\textbf{83.33}$^*$ \\
\mickeyem & \textbf{GRPO} & \textbf{67.04} & \cellcolor{gray!15}\textbf{59.92} & \textbf{77.53} & \textbf{80.25}\\
\bottomrule
\end{tabular}}%
\vspace{-0.1cm}
\end{table}%
}

\newcommand{\pausehinttable}{
\begin{table}[t]
\vspace{-0.1cm}
\centering
\small
\caption{Effect of pause-based visual hints on the blue-arrow subset of DoraVQA
(frames with arrows$\,\geq 1$, Season~2 excluded). The four conditions vary the
spatial signal available at inference: \textbf{No hint} (none),
\textbf{Self-localized} (model's own bbox), \textbf{Prompted hint} (textual
instruction only), \textbf{Oracle box} (ground-truth bbox).
\textit{Loc.}\ reports mean IoU and is only meaningful for \textit{Self-localized}.
\textbf{Bold}: best overall per column. \underline{Underline}: best per model.}
\label{tab:pause_hint}
\resizebox{\columnwidth}{!}{%
\begin{tabular}{@{}llcccccc@{}}
\toprule
\textbf{Model} & \textbf{Condition}
  & \textbf{Overall} & \textbf{Spatial} & \textbf{Nav.}
  & \textbf{Know.} & \textbf{Prob.\ solving} & \textbf{Loc.\ (mIoU)} \\
\midrule
\multirow{4}{*}{Qwen3-VL-8B}
  & No hint        & 52.24 & 58.17 & 35.38 & 56.10 & 66.67 & --   \\
  & Self-localized & 53.61 & 55.83 & 39.52 & \underline{64.86} & 73.61 & 0.76 \\
  & Prompted hint  & \underline{55.94} & \underline{64.59} & 31.54 & 57.14 & \underline{79.17} & --   \\
  & Oracle box     & 55.25 & 61.24 & \underline{45.60} & 46.15 & 67.61 & --   \\
\midrule
\multirow{4}{*}{Qwen3-VL-8B + GRPO}
  & No hint        & 63.19 & 70.48 & 34.62 & 52.38 & 86.11 & --   \\
  & Self-localized & 66.98 & \textbf{80.00} & 36.92 & \textbf{69.05} & 83.33 & \textbf{2.57} \\
  & Prompted hint  & \textbf{67.61} & 77.14 & \textbf{46.92} & 52.38 & 84.72 & --   \\
  & Oracle box     & \textbf{67.61} & 77.62 & 43.85 & 64.29 & \textbf{87.50} & --   \\
\bottomrule
\end{tabular}}
\vspace{-0.1cm}
\end{table}
}

\newcommand{\ablationtable}{%
\begin{table}[t]
\vspace{-0.1cm}
\centering
\caption{Ablation study. \textbf{(a)}~Input modality ablation over visual frames~(V), transcript~(T), and pedagogical context~(C). \textbf{(b)}~Pedagogical structure ablation with all modalities present. A vertical rule separates in-domain benchmarks (DoraVQA, ClubHVQA) from out-of-domain ones (V-MME and NExT-QA). Best results per section in \textbf{bold}. $\dagger$Shared full-model row.}
\label{tab:ablation}
\vspace{4pt}
\resizebox{\textwidth}{!}{%
\begin{tabular}{@{}ccc|cc|cc@{\hskip 14pt}|@{\hskip 14pt}l|cc|cc@{}}
\toprule
\multicolumn{7}{c}{\textbf{(a) Modality Ablation}}
& \multicolumn{5}{c}{\textbf{(b) Structure Ablation}} \\
\cmidrule(r){1-7} \cmidrule(l){8-12}
V & T & C & \doraem & \mickeyem & V-MME & NExT
& Structure & \doraem & \mickeyem & V-MME & NExT \\
\midrule
\cmark & \xmark & \xmark & 43.42 & 26.34 & 60.74 & 67.91
& Random Clips & 49.81 & 41.63 & 51.70 & 70.08    \\
\xmark & \cmark & \xmark & 50.95 & 31.88 & 60.00 & 70.48
& No Pause     & 45.31 & 39.43 & 51.11 & 72.00    \\
\cmark & \cmark & \xmark & 54.72 & 39.77 & 61.11 & 70.83
& & & & & \\
\midrule
\cmark & \cmark & \cmark & \textbf{55.11} & \textbf{54.45} & \textbf{62.20} & \textbf{72.86}
& Full$^\dagger$ & \textbf{55.11} & \textbf{54.45} & \textbf{62.20} & \textbf{72.86} \\
\bottomrule
\end{tabular}%
}%
\end{table}%
\vspace{-0.1cm}
}

\newcommand{\contaminationtable}{
\begin{table}[t]
\centering
\small
\caption{Combined contamination analysis across \doraem~DoraVQA and \mickeyem~ClubHVQA.
The \textbf{LLM backbone} columns (vision encoder disabled) confirm pre-training
contamination via two probes: \textit{MCQ} accuracy well above the 25\% chance
level on episode-specific questions, and \textit{F1} / \textit{Containment}
on open-ended recall (Containment measures whether the gold answer appears
anywhere in the model's output, regardless of phrasing).
The \textbf{Full VLM} columns show this textual knowledge does not transfer to
visual grounding: \textit{Blank image} accuracy remains near chance despite the
model knowing the answers textually, and the gap to \textit{Episode frame (+ GRPO)}
confirms our gains reflect structured visual learning rather than memorization.}
\label{tab:contamination}
\resizebox{\textwidth}{!}{%
\begin{tabular}{@{}l cc cc cc cc cc cc@{}}
\toprule
& \multicolumn{6}{c}{\textbf{LLM backbone (text only)}}
& \multicolumn{6}{c}{\textbf{Full VLM}} \\
\cmidrule(lr){2-7}\cmidrule(lr){8-13}
& \multicolumn{2}{c}{\textbf{MCQ}}
& \multicolumn{2}{c}{\textbf{F1}}
& \multicolumn{2}{c}{\textbf{Containment}}
& \multicolumn{2}{c}{\textbf{Blank image}}
& \multicolumn{2}{c}{\textbf{Episode frame (base)}}
& \multicolumn{2}{c}{\textbf{Episode frame (+ GRPO)}} \\
\cmidrule(lr){2-3}\cmidrule(lr){4-5}\cmidrule(lr){6-7}\cmidrule(lr){8-9}\cmidrule(lr){10-11}\cmidrule(lr){12-13}
\textbf{Model}
  & \doraem & \mickeyem
  & \doraem & \mickeyem
  & \doraem & \mickeyem
  & \doraem & \mickeyem
  & \doraem & \mickeyem
  & \doraem & \mickeyem \\
& \multicolumn{2}{c}{\textit{\small chance = 25\%}}
& \multicolumn{2}{c}{\textit{\small chance $\approx$ 0\%}}
& \multicolumn{2}{c}{\textit{\small chance $\approx$ 0\%}}
& \multicolumn{2}{c}{\textit{\small chance = 25\%}}
& \multicolumn{2}{c}{\textit{\small chance = 25\%}}
& \multicolumn{2}{c}{\textit{\small chance = 25\%}} \\
\midrule
Qwen2-VL-2B  & 34.60 & 19.20 & 5.23  & 4.32  & 22.01 & 19.14 & 33.20 & 31.35 & 41.36 & 31.43 & 55.11 & 36.68 \\
Qwen2-VL-7B  & 36.37 & 21.00 & 5.46  & 9.49  & 23.74 & \textbf{23.65} & 39.49 & 38.63 & 56.74 & 45.09 & 62.38 & 50.10 \\
Qwen3-VL-8B  & \textbf{39.23} & \textbf{27.38} & \textbf{12.07} & \textbf{11.57} & \textbf{24.64} & 16.22 & \textbf{44.54} & \textbf{48.87} & \textbf{58.08} & \textbf{50.45} & \textbf{67.98} & \textbf{56.81} \\
\bottomrule
\end{tabular}}
\end{table}
}
\newcommand{\ablationhyperparameters}{
\begin{table}[H]
\centering
\caption{Sequential hyperparameter sweeps for GRPO training. Each sweep fixes previously optimized parameters. Bold indicates selected values used in main experiments.}
\label{tab:hyperparam_sweep}
\small
\begin{tabular}{@{}lcccc@{}}
\toprule
\textbf{Parameter} & \textbf{Values Tested} & \textbf{Selected} & \textbf{DoraVQA} \\
\midrule

KL Coefficient 
& \{0.0, 0.001, \textbf{0.01}, 0.1\}
& \textbf{0.01}
& \{42.96, 40.71, \textbf{46.38}, 45.19\} \\

Learning Rate 
& \{1e-6, 1e-5, \textbf{1e-4}, 1e-3\}
& \textbf{1e-4}
& \{43.69, 45.19, \textbf{50.09}, 38.89\} \\

Reward Scaling 
& \{0.5, 1.0, \textbf{2.0}, 5.0\}
& \textbf{2.0}
& \{45.92, 45.19, \textbf{45.69}, 43.93\} \\

\bottomrule
\end{tabular}
\end{table}
}

\newcommand{\ablationtrainingsteps}{
\begin{table}[H]
\centering
\caption{Ablation study on training steps for both models. Performance peaks at between 150 and 250 steps then degrades due to overfitting on the narrow educational domain.}
\label{tab:ablation_steps}
\small
\begin{tabular}{@{}lcccccc@{}}
\toprule
\textbf{Model} & 50 & 100 & 150 & 200 & 250 & 300 \\
\midrule
Qwen2-VL-7B 
& 57.49 
& 60.29 
& \textbf{62.96} 
& 60.98 
& 56.00 
& 60.84 \\

Qwen3-VL-2B
& 63.67
& 66.67
& 65.91
& 65.54
& \textbf{67.97}
& 66.60 \\
\bottomrule
\end{tabular}
\end{table}
}

\newcommand{\pausehintlargetable}{
\begin{table}[H]
\centering
\small
\caption{Pause-based visual hint results across model scales (blue-arrow subset).
Same setup and conditions as Table~\ref{tab:pause_hint}.
Qwen3-VL-8B + GRPO (8B) surpasses both Qwen3-VL-30B and Qwen2-VL-72B on overall
accuracy, while Qwen2.5-VL-72B is the only untuned model with meaningful
localization (11.87\% mIoU).
The \textit{self-localized} condition degrades Qwen3-VL-30B sharply
(35.19 vs.\ 55.52 overall), revealing that the coordinate-emission pathway is
unstable at this scale---whereas the \textit{prompted hint}, which carries no
coordinates, is robust across all model sizes.
\textbf{Bold}: best overall per column. \underline{Underline}: best per model.}
\label{tab:pause_hint_large}
\begin{tabular}{@{}llccccc@{}}
\toprule
\textbf{Model} & \textbf{Condition}
  & \textbf{Overall} & \textbf{Spatial} & \textbf{Nav.}
  & \textbf{Know.} & \textbf{Loc.\ (mIoU)} \\
\midrule
\multirow{4}{*}{Qwen3-VL-8B}
  & No hint        & 52.24 & 58.17 & 35.38 & 56.10 & --   \\
  & Self-localized & 53.61 & 55.83 & 39.52 & \underline{64.86} & 0.76 \\
  & Prompted hint  & \underline{55.94} & \underline{64.59} & 31.54 & 57.14 & --   \\
  & Oracle box     & 55.25 & 61.24 & \underline{45.60} & 46.15 & --   \\
\midrule
\multirow{4}{*}{Qwen3-VL-8B + GRPO}
  & No hint        & 63.19 & 70.48 & 34.62 & 52.38 & --   \\
  & Self-localized & 66.98 & \textbf{80.00} & 36.92 & \textbf{69.05} & 2.57 \\
  & Prompted hint  & \textbf{67.61} & 77.14 & \textbf{46.92} & 52.38 & --   \\
  & Oracle box     & \textbf{67.61} & 77.62 & 43.85 & 64.29 & --   \\
\midrule
\multirow{4}{*}{Qwen3-VL-30B}
  & No hint        & \underline{55.52} & \underline{63.24} & 25.81 & 48.78 & --   \\
  & Self-localized & 35.19 & 39.80 & 19.49 & 39.47 & 0.00 \\
  & Prompted hint  & 54.37 & 63.05 & 28.00 & 51.28 & --   \\
  & Oracle box     & 52.69 & 59.90 & \underline{30.17} & \underline{61.11} & --   \\
\midrule
\multirow{4}{*}{Qwen2-VL-72B}
  & No hint        & 52.85 & 57.62 & 31.78 & 52.38 & --   \\
  & Self-localized & \underline{56.24} & 63.81 & \underline{36.15} & 50.00 & 2.03 \\
  & Prompted hint  & 55.45 & \underline{64.29} & 33.85 & \underline{64.29} & --   \\
  & Oracle box     & 55.45 & 61.90 & 33.85 & 59.52 & --   \\
\midrule
\multirow{4}{*}{Qwen2.5-VL-72B}
  & No hint        & 56.56 & 61.90 & 40.00 & 50.00 & --    \\
  & Self-localized & 54.19 & 56.19 & 34.62 & 61.90 & \textbf{11.87} \\
  & Prompted hint  & \underline{60.66} & \underline{66.19} & 40.77 & 59.52 & --    \\
  & Oracle box     & 58.77 & 65.71 & \underline{42.31} & \underline{64.29} & --    \\
\bottomrule
\end{tabular}
\end{table}
}

\newcommand{\localizationtable}{
\begin{table}[H]
\centering
\small
\caption{Arrow localization accuracy across model scales (\textit{self-localized}
condition only, where coordinates are externalized).
IoU@$k$ reports the percentage of predictions exceeding the given threshold.
The GRPO-finetuned 8B model (2.57\% mIoU) outperforms both the 30B and 72B
untuned models, demonstrating that structured supervision recovers spatial
localization more effectively than scaling alone.}
\label{tab:arrow_localization}
\begin{tabular}{@{}lccccc@{}}
\toprule
\textbf{Model} & \textbf{IoU@10} & \textbf{IoU@25} & \textbf{IoU@50} & \textbf{IoU@75} & \textbf{mIoU} \\
\midrule
Qwen3-VL-8B           & 0.00    & 0.00    & $<$1  & $<$1  & 0.76  \\
Qwen3-VL-8B + GRPO    & 0.00    & 0.00    & $<$1  & $<$1  & 2.57  \\
Qwen3-VL-30B          & 0.00  & 0.00  & 0.00  & 0.00  & 0.00  \\
Qwen2-VL-72B          & 6.96  & 0.63  & 0.00  & 0.00  & 2.03  \\
Qwen2.5-VL-72B        & \textbf{35.51} & \textbf{20.22} & \textbf{5.10}  & \textbf{0.80}  & \textbf{11.87} \\
\bottomrule
\end{tabular}
\end{table}
}

\title{Structure Over Scale:\\Learning Visual Reasoning from Pedagogical Video}

\author{%
  Bishoy Galoaa* \\
  Northeastern University \\
  \texttt{galoaa.b@northeastern.edu}
  \And
  Xiangyu Bai* \\
  Northeastern University \\
  \texttt{bai.xiang@northeastern.edu}
  \And
  Sarah Ostadabbas \\
  Northeastern University \\
  \texttt{s.ostadabbas@northeastern.edu}
}

\begin{document}

\maketitle

\begin{abstract}
State-of-the-art vision-language models (VLMs) score impressively on video
benchmarks yet stumble on basic visual reasoning tasks involving spatial
relations, navigation, and object selection that a preschooler solves
easily. We hypothesize that the explicit pedagogical structure,
specifically the \textit{context-question-pause-answer} cycles embedded in
children's educational video, provides naturally co-aligned reasoning
traces: temporally synchronized visual cues, questions, and answers that
emerge only from deliberate pedagogical authoring and cannot be practically
reconstructed through manual annotation at scale. To test this, we introduce
\textbf{SoSVQA} (Structure over Scale Visual Question Answering), a unified
benchmark of 10K question-answer pairs automatically extracted from
\textit{Dora the Explorer} (DoraVQA) and \textit{Mickey Mouse Clubhouse}
(ClubHVQA) with precise timestamp alignment, and fine-tune Qwen2-VL and
Qwen3-VL using Group Relative Policy Optimization (GRPO) to leverage the
clear correctness signals and structured reasoning traces inherent in
educational content. Despite training on just 10K QA pairs from 78 hours
of children's television, orders of magnitude less data than GPT and Gemini, our approach delivers generalizable performance gains for Qwen-based VLMs, yielding consistent improvements on NExT-QA (+19.7), Video-MME
(+10.6), and MotionBench (+4.9), matching the performance of leading proprietary systems and demonstrating that content structure can compensate for
content scale. Code is available at \href{https://github.com/ostadabbas/SoS-Structure-Over-Scale}{\faGithub\ \texttt{ostadabbas/SoS}}.
\end{abstract}

\vspace{-0.1cm}
\section{Introduction}
\vspace{-0.1cm}

Despite achieving strong performance~\cite{xu2017videoquestion, xiao2021next}, today's vision-language models fail systematically at the simple reasoning tasks that children master before kindergarten. On diagnostic benchmarks targeting objects, spatial relations, and directional understanding, state-of-the-art models such as GPT-5.5 and Gemini 3.1 Pro achieve only 60--70\% accuracy \cite{singh2025openai, deepmind_gemini_3_1_pro_2026}, far below the 95\%+ performance observed in humans~\cite{fu2024blink}. These failures persist even in simple scenes~\cite{tong2024eyes}, suggesting that gains in video understanding often arise from scaling data volume and statistical pattern matching rather than grounded reasoning. Modern data-centric models largely memorize surface level patterns: state-of-the-art VLMs display only superficial reasoning, mirroring patterns from large-scale, generic video-text pairs rather than genuinely learning spatial relational reasoning. We argue that effective learning of spatial concepts requires not only large data corpus, but also repeated, explicit supervision that grounds language to visual evidence through instruction, feedback, and temporal scaffolding.

Children's educational television offers this uniquely structured alternative. Programs such as \textit{Dora the Explorer}~\cite{dora_explorer} and \textit{Mickey Mouse Clubhouse}~\cite{gannaway2006mmclubhouse} are explicitly designed around a consistent \textit{context-question-pause-answer} pedagogical loop: each episode establishes visual context, poses direct spatial questions to the viewer, and then pauses (visually ``hinting'' at the answer by overlaying directional arrows, zooms, and character gestures that isolate the exact evidence needed to answer) before revealing an unambiguous answer with verbal explanation. This pause segment functions as an implicit reasoning trace, encoding two distinct learning signals that we later isolate experimentally (Section~\ref{sec:pause_hint}): a spatial attention signal (where to look, recoverable through the visual hints) and an entity grounding signal (what it is called). The structure recurs systematically across every entry, generating 15--20 explicit question-answer pairs with clear correctness signals per episode.

\sosvqafigure

This pedagogical design produces measurable learning gains in children \cite{linebarger2005vocabulary, anderson2000blueclues, angin2017spatial}, with controlled studies showing that preschoolers who watch \textit{Dora the Explorer} score significantly higher on direction/position concepts and spatial ability tasks than children who do not \cite{angin2017spatial}, confirming that the pause structure carries genuine supervisory signal. We hypothesize that the same structure that supports human learning can serve as an efficient and underexplored training signal for vision language models. This observation points to a broader scalable data strategy: rather than blindly rely on increasing data scale; we need data that knows how to \textit{teach}. Fortunately, the world contains a large supply of pedagogically structured content that has never been used this way: children's television, Khan Academy lectures, PhET simulations, and similar resources all embed explicit teaching signals in their design.  To test this hypothesis, we introduce \textbf{SoSVQA} (Structure over Scale Visual Question Answering), a unified benchmark of 10K question-answer pairs, containing two sub-datasets extracted from children's educational television: \textbf{DoraVQA}, comprising 5{,}348 question-answer pairs from 8 seasons (96 episodes) of \textit{Dora the Explorer}, and \textbf{ClubHVQA} (ClubhouseVQA), comprising 4{,}652 question-answer pairs from 4 seasons (132 episodes) of \textit{Mickey Mouse Clubhouse}. Each example preserves the show's pedagogical structure (context, question, pause, and answer) with corresponding video timestamps and pause-based visual context extracted via an LLM agent.

To test whether structured supervision transfers beyond its source content, we fine-tune Qwen2-VL~\cite{wang2024qwen2vl} and Qwen3-VL~\cite{qwen3vl2025} using Group Relative Policy Optimization (GRPO)~\cite{shao2024deepseekmath}, which naturally aligns with educational content: answers are objectively correct or incorrect, and the show's explanations provide implicit reasoning cues without manual annotation. Rather than adapting models to specific content, our goal is to improve spatial reasoning abilities using structured learning format already embedded in the data. We evaluate along three axes: (i) \textbf{in-domain} performance on DoraVQA and ClubHVQA, (ii) \textbf{cross-show} transfer between the two subsets to test whether learning generalizes across pedagogical content rather than memorizing show-specific patterns, and (iii) \textbf{cross-benchmark} transfer to Video-MME~\cite{fu2025video}, NExT-QA~\cite{xiao2021next}, MVBench~\cite{li2024mvbench}, MotionBench~\cite{hong2025motionbench}, and TVBench~\cite{cores2024tvbench}. We train on open-ended answer generation but evaluate on multiple-choice benchmarks throughout, a deliberate format mismatch that distinguishes transferable reasoning from answer-format memorization. We additionally probe whether the show's \textit{pause hints} are recoverable by prompting alone or require structured supervision (Section~\ref{sec:pause_hint}), separating spatial attention from entity grounding as distinct learning signals.


Our contribution can be listed as follows:
\begin{itemize}
    \item We show that the \textit{context-question-pause-answer} structure
    of children's educational television serves as implicit reasoning
    supervision for VLMs, yielding cross-benchmark spatial reasoning
    improvements without large-scale data.
    
    \item We present an automated pipeline that extracts self-supervised
    training data from pedagogically structured video, preserving the
    correctness signals embedded in its reasoning structure without manual
    annotation. We validate the pipeline across two shows with distinct
    reasoning profiles, and additionally release manually annotated
    bounding-box hints for the pause-segment visual cues to support future
    spatial-grounding work.

    \item We introduce \textbf{SoSVQA}, a benchmark of 10K question-answer
    pairs automatically extracted from \textit{Dora the Explorer} and \textit{Mickey Mouse Clubhouse} with precise timestamp alignment. SoSVQA is the first video question-answering datasets extracted
    from children's educational television, with temporal alignment between
    visual frames, questions, context, and answers.
\end{itemize}

\vspace{-0.1cm}
\section{Related Work}
\vspace{-0.1cm}
\textbf{VLMs and Video Understanding.}
Recent work has scaled VLMs to capture temporal dynamics beyond images. Video-LLaVA~\cite{lin2024videollava} demonstrated that joint training on videos and images improves both modalities through unified visual representations; InternVideo2~\cite{wang2024internvideo2} expanded video encoders using progressive training and masked video modeling; LLaVA-Video~\cite{zhang2024llavavideo} introduced dense frame sampling to capture fine-grained temporal dynamics. However, these approaches train predominantly on web videos~\cite{miech2019howto100m, abu2016youtube, kay2017kinetics}, which contain diverse content but lack explicit pedagogical structure. While these videos display spatial relationships and temporal sequences, they rarely verbalize spatial concepts explicitly or provide structured question-answer pairs that ground visual patterns in linguistic descriptions. Our work differs by leveraging educational content where spatial reasoning is explicitly taught through scaffolded question-answer sequences with clear correctness signals.

\textbf{Spatial Reasoning in VLMs.}
Despite achieving strong performance on existing video benchmarks~\cite{xiao2021next, xu2017videoquestion}, VLMs exhibit systematic failures on spatial reasoning tasks. Fu et al.~\cite{fu2024blink} reveals that VLMs achieve only half of human accuracy with noticeable failures on spatial reasoning tasks. Tong et al.~\cite{tong2024eyes} demonstrated that current VLMs are barely above random chance on tasks requiring precise spatial understanding. These failures extend to fundamental spatial primitives: models struggle with counting objects beyond small numbers~\cite{paiss2023teaching}, fail at compositional spatial relationships~\cite{thrush2022winoground}, and fail to reason about object positions across video frames~\cite{shangguan2024tomato}. Importantly, these deficiencies persist even in simple visual scenes with minimal occlusion and clear object boundaries, indicating that the limitation is not perceptual but conceptual: models lack the structured spatial reasoning that children acquire through explicit instruction.

\textbf{Educational and Instructional Video as Training Data.}
Instructional videos have been explored as training data for procedure understanding and action recognition. Miech et al.~\cite{miech2019howto100m} demonstrated that instructional videos provide natural language supervision for video-text embeddings. However, the narrations describe actions chain rather than teaching concepts; The CrossTask~\cite{zhukov2019crosstask} and COIN~\cite{tang2019coin} datasets extract procedural steps from instructional videos but focus on temporal ordering rather than spatial reasoning. Although these datasets compose of structured contents, they are not explicit question-answer structure that ``teaches'' spatial reasoning. Closest to our work, Anderson and Burns~\cite{anderson2000blueclues} analyzed the pedagogical structure of children's educational television, documenting how the question-pause-answer format creates structured learning opportunities. We build on this insight by extracting the explicit question-pause-answer pairs from educational content as direct supervision signals, leveraging the pedagogical design for model training rather than treating video as passive observation. Adjacent work has trained VLMs on infant egocentric video~\cite{wang2025babyvlm}, but no prior work has used the pedagogical structure of children's television as a direct training signal for VLMs.
\doravqapipeline

\textbf{Reinforcement Learning for VLM Fine-tuning.}
Group Relative Policy Optimization (GRPO)~\cite{shao2024deepseekmath} has emerged as an efficient alternative to Proximal policy optimization (PPO) \cite{schulman2017proximal} for reasoning tasks:  DeepSeek-R1~\cite{deepseek2025r1} demonstrated that GRPO with rule-based rewards achieves reasoning capabilities competitive with proprietary models. Recent work has successfully applied GRPO to vision-language tasks: VLM-R1~\cite{shen2025vlmr1} showed that GRPO fine-tuning achieves superior out-of-distribution generalization compared to SFT on visual reasoning tasks, while CrowdVLM-R1~\cite{wang2025crowdvlm} developed specialized reward functions for counting tasks in crowded scenes. These works primarily design reward models for general-purpose instruction following. In contrast, we apply GRPO to pedagogically-structured video content where the reward signal is implicit in the educational design, where ground-truth answers are directly in its structure. This eliminates the need for external reward model while providing clear correctness signals, creating a self-supervised RL training environment where the pedagogical content naturally defines the learning objectives.

Our work is positioned at the intersection of these research directions: we address video-language model spatial reasoning failures by training on content that explicitly teaches spatial concepts through structured question-answer pairs. Unlike general instructional videos that show only procedures, educational television teaches reasoning through direct instruction, repetition, and scaffolded examples. We leverage this pedagogical structure with GRPO fine-tuning, using the natural correctness signals from educational content as supervision without requiring manual annotation or reward engineering.
\vspace{-0.1cm}
\section{Method}

\vspace{-0.1cm}
\subsection{SoSVQA: Structure and Formatting}
\vspace{-0.1cm}

We operationalize the \textit{context-question-pause-answer} structure of educational television as a multimodal training example (Figure~\ref{fig:doravqa-pipeline}). Each example in the SoSVQA dataset $\mathcal{D}$ is formatted as $x = \{I, T, Q, a^*\} \in \mathcal{D}$, where $I$ represents the visual frames sampled before and at the question timestamp, $T$ is the transcript providing narrative context within a fixed time window before the question, $Q$ is the explicit question and $a^*$ is ground truth answer. We treat $I$ and $T$ as separate input streams rather than collapsing them, allowing the model to draw on visual scene evidence and linguistic context independently before fusion. The ground-truth answer $a^*$ is always either directly stated in the transcript or deducible in a single inferential step, providing an unambiguous reward target without manual annotation. The model receives $x$ as a structured multimodal prompt (visual frames embedded alongside a text sequence concatenating the system instruction, transcript context, and question) and generates a distribution over answer tokens that is trained toward $a^*$ via the reward signal described in Section~\ref{sec:grpo}.

\vspace{-0.1cm}
\subsection{Group Relative Policy Optimization (GRPO)}
\vspace{-0.1cm}
\label{sec:grpo}

We fine-tune using GRPO~\cite{shao2024deepseekmath}, which optimizes a policy 
$\pi_\theta$ without a separate value network by computing advantages relative 
to a group of sampled outputs. GRPO is well-suited here because educational 
content provides objectively correct answers, eliminating the need for complex 
reward modeling. For each input $x \in \mathcal{D}$, the policy samples $K$ candidate answers 
$\{a_1, \ldots, a_K\}$. Each candidate is scored against the ground-truth 
$a^*$ via:
\begin{equation}
    r(a_i, a^*) = \alpha\,\text{F1}(a_i, a^*) + \beta \left(1 - 
    \frac{\text{lev}(a_i, a^*)}{\max(|a_i|, |a^*|)}\right),
\label{eq:rewardf}
\end{equation}
where $\text{F1}(a_i, a^*)$ measures token-level semantic overlap, 
$\text{lev}(a_i, a^*)$ is the Levenshtein edit distance normalized by the 
length of the longer string $\max(|a_i|, |a^*|)$, and $\alpha = 0.3$, 
$\beta = 0.7$ are fixed weights selected by empirical validation (see Appendix \ref{sec:hyperparam_sweeps}), with 
$\alpha + \beta = 1$. The two terms are complementary: F1 tolerates minor 
surface variation while normalized Levenshtein penalizes character-level 
errors and near-miss re-orderings. This yields a stable continuous signal; 
exact matches score $1.0$, semantically correct answers with minor variation 
score $0.65$--$0.95$, and incorrect answers score near zero.

For a group of $K$ sampled candidates with rewards 
$\{r(a_1, a^*), \ldots, r(a_K, a^*)\}$, the advantage for candidate $a_i$ 
is computed relative to the group mean:
\begin{equation}
    A(a_i, a^*) = r(a_i, a^*) - \frac{1}{K}\sum_{j=1}^{K} r(a_j, a^*),
\end{equation}
so that the policy is updated toward outputs that outperform the group 
average rather than toward an absolute value target. The training objective 
is:
\begin{equation}
    \mathcal{L}(\theta) = -\mathbb{E}_{x \sim \mathcal{D},\, 
    a_i \sim \pi_\theta(\cdot|x)}\left[A(a_i, a^*)\log\pi_\theta(a_i|x)
    \right].
\end{equation}

Critically, the reward in Eq.~\ref{eq:rewardf} is computed against free-form
$a^*$ during training, while evaluation uses multiple-choice prompting.
This deliberate format mismatch tests whether gains reflect transferable
reasoning rather than answer-format memorization.

\vspace{-0.1cm}
\subsection{GRPO Training}
\vspace{-0.1cm}
\label{sec:training}

The GRPO training procedure for the \textit{context-question-pause-answer}  structure is presented in Algorithm~\ref{alg:grpo-training} and illustrated in Figure~\ref{fig:doravqa-pipeline}. For each multimodal input $x \in \mathcal{D}$, the policy $\pi_\theta$ generates 
$K$ candidate answers $\{a_1, \ldots, a_K\}$ via temperature sampling 
(temperature $= 1.0$, top-$p = 0.9$). Each candidate $a_i$ is evaluated 
against the ground-truth answer $a^*$ using the reward function 
$r(a_i, a^*)$ defined in Section~\ref{sec:grpo}, and group-relative 
advantages $A(a_i, a^*)$ are computed over the $K$ generations. The policy parameters $\theta$ are updated by minimizing $\mathcal{L}(\theta)$ to 
maximize the expected group-relative advantage. To stabilize training, a 
KL divergence penalty is added between the updated policy $\pi_\theta$ and 
a frozen reference policy $\pi_{\text{ref}}$:
\begin{equation}
    \mathcal{L}_{\text{KL}}(\theta) = \mathcal{L}(\theta) + 
    \lambda\, \mathbb{E}_{x \sim \mathcal{D}}\left[
    D_{\text{KL}}\!\left(\pi_\theta(\cdot|x) \,\|\, 
    \pi_{\text{ref}}(\cdot|x)\right)\right],
\end{equation}
where $\lambda = 0.01$ is the KL coefficient. Unless otherwise noted, models are trained jointly on both SoSVQA splits; single-show training is used only in the cross-show ablation (Table~\ref{tab:ablation_method}). Training hyperparameters (learning rate $\eta = 1 \times 10^{-4}$, reward scaling factor $s = 2.0$, and group size $K = 8$) are selected via sequential sweeps on held-out validation data (Appendix~\ref{sec:hyperparam_sweeps}).

\begin{algorithm}[t]
\caption{GRPO Training with Context-Question-Pause-Answer Structure}
\label{alg:grpo-training}
\begin{algorithmic}[1]
\Require Dataset $\mathcal{D}$ of $x{=}\{I, T, Q, a^*\}$; policy $\pi_\theta$;
         reference $\pi_{\text{ref}}$; group size $K$; weights $\alpha,\beta,\lambda$;
         learning rate $\eta$
\For{each mini-batch $\mathcal{B} \sim \mathcal{D}$}
    \ForAll{$x \in \mathcal{B}$}
        \State Sample $\{a_i\}_{i=1}^{K} \sim \pi_\theta(\cdot \mid I, T, Q)$
        \State $r_i \gets \alpha\,\text{F1}(a_i, a^*) + \beta\!\left(1 - \tfrac{\text{lev}(a_i, a^*)}{\max(|a_i|,|a^*|)}\right)$
        \State $A_i \gets r_i - \tfrac{1}{K}\textstyle\sum_{j} r_j$
    \EndFor
    \State $\theta \gets \theta - \eta\,\nabla_\theta \sum_{x,i}\!\left[-A_i \log\pi_\theta(a_i \mid x) + \lambda\, D_{\text{KL}}(\pi_\theta \,\|\, \pi_{\text{ref}})\right]$
\EndFor
\end{algorithmic}
\end{algorithm}

\quality
\vspace{-0.1cm}
\section{Experiments and Results}
\vspace{-0.1cm}
We evaluate our approach along three axes designed to test the structure-over-scale hypothesis. First, \textit{in-domain} performance on both SoSVQA splits (DoraVQA and ClubHVQA) tests whether the \textit{context-question-pause-answer} structure was learned at all, and contamination analysis verifies that improvements come from visual grounding rather than pre-training memorization. Second, \textit{cross-show} transfer between DoraVQA and ClubHVQA tests whether learning generalizes across pedagogical content rather than memorizing show-specific patterns. Third, \textit{cross-benchmark} transfer to six external video-understanding benchmarks tests whether the learning generalizes beyond children's educational content, and ablations isolate which components drive the gains.

\vspace{-0.1cm}
\subsection{Datasets}
\vspace{-0.1cm}
We evaluate on in-domain SoSVQA (DoraVQA and ClubHVQA) and a broad set of out-of-domain external benchmarks selected to test the transferability of spatial reasoning improvements. Our models achieve strong gains across all benchmarks, ruling out narrow overfitting to a single evaluation style. 

\noindent\textbf{SoSVQA.} The two SoSVQA subsets contain a combined total of 10,000 QA pairs and follow a consistent
\textit{context--question--pause--answer} pedagogical structure with clear
ground-truth supervision, but exhibit complementary reasoning profiles
(Figure~\ref{fig:sosvqa-overview}): DoraVQA (5,348 pairs) is spatial-heavy (60.6\%) with
strong navigation demands, while ClubHVQA (4,652 pairs) is balanced between spatial and
non-spatial tasks (49\%/51\%) and dominated by object selection (78\%).
We do not redistribute video frames or audio; we release only episode
identifiers, timestamps, QA annotations, the annotation pipeline, and
manually annotated bounding-boxes for the pause-segment visual cues.
Multiple-choice options are generated from the free-form gold answers via
Gemini 2.5 Flash and manually audited to prevent surface-form shortcuts
(Appendix~\ref{sec:mcq_audit}).

\doravqamainresults
\noindent\textbf{Video-MME}~\cite{fu2025video} evaluates multimodal LLMs across
short, medium, and long videos using visual, subtitle, and audio cues; we
report the short partition. \textbf{NExT-QA}~\cite{xiao2021next}
tests causal and temporal reasoning through ``why'' and ``how'' questions.
\textbf{MVBench}~\cite{li2024mvbench} covers 20 temporal video-understanding
tasks designed to require multi-frame reasoning. \textbf{MotionBench}~\cite{hong2025motionbench}
evaluates fine-grained motion comprehension through six motion-oriented
question categories. \textbf{TVBench}~\cite{cores2024tvbench} isolates
temporal reasoning by removing the spatial, textual, and world-knowledge
shortcuts present in earlier benchmarks.

\vspace{-0.1cm}
\subsection{In-Domain Performance on SoSVQA}
\vspace{-0.1cm}

Table~\ref{tab:doravqa_main} reports per-category performance on both
DoraVQA and ClubHVQA, testing whether the
\textit{context-question-pause-answer} structure was learned.
GRPO fine-tuning produces large, structurally-aligned gains on the two
categories most directly targeted by the show's pedagogical pause: spatial
reasoning improves by +11 to +13 points across all DoraVQA rows, and
navigation gains as much as +23.25 (Qwen2-VL-2B) and +15.29 (Qwen3-VL-8B)
on ClubHVQA, reflecting the dataset's visually-driven structure. Object
selection and counting improve consistently across subsets (+6 to +13
points), and knowledge recall transfers strongly, peaking at +18.48 for
Qwen2-VL-7B on DoraVQA. These per-category gains aggregate into consistent overall improvements of +5.47 to +8.03 on DoraVQA and +1.57 to +5.25 on ClubHVQA, with larger relative gains on smaller models. The lone regression is Qwen3-VL-8B's navigation on DoraVQA (-0.63), where the baseline already operates near saturation. Figure~\ref{fig:doravqa-qualitative} shows representative qualitative improvements.

\vspace{-0.1cm}
\subsection{Cross-Benchmark Transfer}
\vspace{-0.1cm}
\crossbenchmarkresults
Table~\ref{tab:cross_benchmark} provides empirical support for our structure-over-scale hypothesis: structured pedagogical training induces reasoning capabilities that generalize beyond the narrow domain of children's educational content. With only 10K QA pairs (78 hours), our GRPO-finetuned models match or exceed proprietary systems trained on orders of magnitude more data on multiple external benchmarks. Crucially, all external evaluation uses multiple-choice prompting while training was open-ended, ruling out answer-format memorization as the source of these gains.

The largest improvements appear on \textbf{NExT-QA}~\cite{xiao2021next},
where all three GRPO-finetuned backbones gain +12.46 to +19.70 points,
followed by \textbf{Video-MME} (+1.75 to +10.64). Both benchmarks focus on causal and temporal why/how questions that require integrating evidence across temporally separated frames. Our structured context-question-pause-answer paradigm is designed to address these problems: the question anchors to earlier context, the pause supplies the relevant visual evidence, and the answer follows.
On \textbf{MVBench}, \textbf{MotionBench}, and \textbf{TVBench}, gains are small or absent, because these benchmarks emphasize fine-grained motion and bias-controlled temporal cues that the pedagogical pause does not directly target. The pattern is consistent: where the benchmark probes a reasoning skill structurally aligned with SoSVQA's pedagogical loop, transfer is large; where it does not, gains shrink. This both supports the structure-over-scale claim and delineates its scope.

\contaminationtable
\vspace{-0.1cm}
\subsection{Contamination Analysis}
\vspace{-0.1cm}
A natural concern is that gains on SoSVQA reflect textual memorization of
existing TV program contents from
pre-training sources, rather than genuine visual grounding.
We probe each LLM backbone directly (vision encoder disabled) with factual
recall and Wikipedia completion, and separately evaluate the full VLM with a
blank gray image substituted for every episode frame.
Table~\ref{tab:contamination} shows DoraVQA is markedly more contaminated than
ClubHVQA on every backbone probe. This is unsurprising given Dora's verbally
distinctive format (``to get to X, we go past Y'') generates highly quotable
text across blogs, wikis, and parenting forums, while Clubhouse is
visually-driven, dominated by object selection (78\%), and less represented in
early web crawls.
Crucially, this textual knowledge does \emph{not} transfer to visual
performance. Blank-image accuracy remains near chance for both datasets, and
GRPO gains hold even on the more contaminated DoraVQA, indicating our
improvements come from structured visual grounding rather than recall of
pre-learned facts.
\ablationtrainingmethod

\ablationtable
\vspace{-0.1cm}
\subsection{Ablation Studies}
\vspace{-0.1cm}
We conduct ablation studies on training strategies, input modalities, and pedagogical structure using a Qwen2-VL-2B model fine-tuned for 100 epochs on DoraVQA, evaluating performance on 10\% randomly sampled test subsets of Video-MME and NExT-QA.

\noindent\textbf{Training Method.} Table~\ref{tab:ablation_method} compares SFT and GRPO across the \textit{cross-show} 2$\times$2 matrix (DoraVQA and ClubHVQA as both train and eval) plus three external benchmarks. SFT matches GRPO in-domain (54.30 vs.\ 55.11 on DoraVQA) and achieves modest cross-benchmark gains, but collapses 19.83 points below baseline when evaluated on the unseen show ClubHVQA (30.52 vs.\ 50.35), revealing overfitting to show-specific surface patterns. GRPO generalizes in both cross-show directions, but the transfer is asymmetric: DoraVQA$\rightarrow$ClubHVQA improves 4.10 points over baseline, while ClubHVQA$\rightarrow$DoraVQA improves 25.68 points and even exceeds the Dora-trained GRPO on DoraVQA itself (67.04 vs.\ 55.11). The Mickey-trained model also produces our highest cross-benchmark scores on V-MME (77.53). This asymmetry is the cleanest test we have of the structure-versus-content question: if learning were content-driven, training on Dora should help most on Dora; instead, the show whose pedagogical loop is more visually anchored (ClubHVQA, 78\% object selection) produces a stronger model in either direction. The shared \textit{context-question-pause-answer} structure, not show-specific patterns, is what the model learns.

\noindent\textbf{Generic Training Control.} To isolate whether GRPO itself drives the cross-benchmark gains, Table~\ref{tab:ablation_method}(b) trains the same backbones with GRPO on a generic 10K-pair open-ended VQA set (NExT-OE) of matched size. Despite NExT-OE training-set leakage (marked$^*$), SoSVQA scores barely move from baseline (Qwen3-VL-8B: 56.20/50.75 vs.\ 55.24/50.67), and transfer to V-MME is also
weaker than SoSVQA-joint training (74.44 vs.\ 80.00). The gap isolates pedagogical structure, not GRPO, as the source of gains.

\noindent\textbf{Context Modalities.} Table~\ref{tab:ablation}(a) shows the model learns more effectively from transcripts than visual frames (50.95 vs.\ 43.42 on DoraVQA), consistent with the text-vision asymmetry reported in the Qwen2 technical report~\cite{wang2024qwen2vl}. Combining modalities compounds the gain on both shows, with the multimodal jump particularly pronounced on ClubHVQA (V+T: 39.77 vs.\ T-only: 31.88), reflecting its higher proportion of visually-grounded object-selection tasks. 
\pausehinttable

\noindent\textbf{Pedagogical Structure.} Removing the pause segment drops DoraVQA accuracy by 9.80 points, and substituting random (non-pause) clips loses 5.30 points (Table~\ref{tab:ablation}(b)). The temporal alignment between question and pause-segment visual evidence is itself a learning signal: the same frames presented without pause alignment recover only part of the gain. To further isolate the two learning signals encoded in the pause—\textit{spatial attention} (where to look) and \textit{entity grounding} (what it is called)—we evaluate four conditions on a held-out blue-arrow subset of DoraVQA, varying how directly the arrow's location is made available to the model (Table~\ref{tab:pause_hint}). Prompting recovers most of the spatial-attention signal but cannot bridge from location to a named entity; only GRPO supervision closes that gap. The pattern holds across model scales up to 72B (Appendix~\ref{sec:pause_hint}, Tables~\ref{tab:pause_hint_large} and~\ref{tab:arrow_localization}), where larger untuned models partially recover spatial attention through scale alone but never match the GRPO-finetuned 8B on entity grounding.
Finally, we examine how effectively VLMs leverage the visual cues embedded in pause segments--such as the blue arrows in DoraVQA and the glowing outline in ClubHVQA--that subtly guide viewers toward the correct answer. Full details of this analysis are provided in Appendix \ref{sec:pause_hint}.
\vspace{-0.1cm}
\section{Conclusion}
\vspace{-0.1cm}
We demonstrated that the pedagogical structure of children's educational television provides an effective training signal for vision-language reasoning. By fine-tuning Qwen2-VL and Qwen3-VL on just 10K QA pairs from SoSVQA using GRPO, we achieved substantial and generalizable performance gain on various benchmarks,
outperforming models trained on orders of magnitude more data. 
While our current training relies on only two pedagogical video sources, this work establishes a foundational step toward a comprehensive Pedagogical Interactive Structure (PIS) dataset spanning realistic structured sources (Khan Academy, PhET simulations, classroom recordings) across multiple levels and reasoning domains, all while integrating visual reward modeling to capture motion-based teaching signals during the pause, opening a new chapter for spatial reasoning in VLMs.


\bibliographystyle{plainnat}
\bibliography{ref}

\newpage
\appendix
\onecolumn
\section{Hyperparameter Sweeps}
\label{sec:hyperparam_sweeps}

We conduct sequential hyperparameter sweeps to identify optimal GRPO training configurations. Each sweep fixes previously optimized hyperparameters while varying a single dimension.

\ablationhyperparameters
Table~\ref{tab:hyperparam_sweep} presents our hyperparameter optimization results. The KL coefficient controls the divergence penalty between the policy and reference model; we find 0.01 provides the best balance, with both higher (0.1) and lower (0.001) values degrading performance. For learning rate, 1e-4 achieves optimal performance (50.09\%), while higher rates (1e-3) cause training instability (38.89\%) and lower rates (1e-6) result in insufficient learning. Reward scaling at 2.0 provides marginal improvements over the default 1.0, amplifying the learning signal from correct answers without overwhelming the policy gradient.

Table~\ref{tab:ablation_steps} examines training duration and overfitting characteristics. Qwen2-VL-7B peaks at 150 steps (62.96\%) before degrading sharply at 250 steps (56.0\%), indicating overfitting to the narrow educational domain. Qwen3-VL-2B shows more robust training, peaking later at 250 steps (67.97\%) with gradual performance decline. These results highlight the importance of early stopping when fine-tuning on domain-specific content to preserve generalization capabilities.
\ablationtrainingsteps
Based on these sequential sweeps, we use KL coefficient = 0.01, learning rate = 1e-4, and reward scaling = 2.0 for all main experiments reported in this paper. Training duration is set at 150 steps for Qwen2-VL-7B and 250 steps for Qwen3-VL-2B to avoid overfitting while maximizing in-domain performance.

\section{Pause Visual Hint Analysis}
\label{sec:pause_hint}

A central hypothesis of this work is that the \textit{pause} segment in the
\textit{context--question--pause--answer} structure functions as an implicit reasoning
trace: it visually isolates the evidence required to answer the question, thereby
grounding language in perception.
To test whether this signal is recoverable by an off-the-shelf vision-language model,
we design a controlled prompting experiment that makes the show's blue-arrow hint
explicitly available to the model, and measure how much of the resulting gain requires
GRPO fine-tuning versus prompt engineering alone. To enable this analysis and support future spatial-grounding work, we manually annotated 702 bounding boxes around the blue directional arrows~\bluearrow~that appear during \textit{Dora the Explorer} pause segments, released alongside the SoSVQA pipeline.

\paragraph{Setup.}
We evaluate four conditions on a held-out subset of DoraVQA examples that contain
a visible blue directional arrow during the pause segment. The conditions form an
intervention-strength ladder over the spatial signal available at inference:
\begin{itemize}
    \item \textbf{No hint.} Standard inference; the model is not told the arrow exists.
    \item \textbf{Self-localized.} The model emits a bounding box for the arrow in a
    first pass and conditions on its own (noisy) prediction in a second pass.
    \item \textbf{Prompted hint.} The model receives a textual instruction:
    \textit{``Locate the blue arrow in the image. Use its bounding-box coordinates to
    identify the object or region it is pointing at, and use that information to answer
    the question.''}---no coordinates are produced or supplied externally.
    \item \textbf{Oracle box.} The ground-truth bounding box of the arrow is supplied
    directly in the prompt.
\end{itemize}
We additionally evaluate each condition with our \textbf{+ GRPO} fine-tuned model.
All conditions use identical visual frames and question text; the only variables are the
system prompt and whether the model weights have been fine-tuned.

\paragraph{Qualitative findings.}
Figure~\ref{fig:pause-hint-qualitative} illustrates two representative examples.
In \textit{Example~1} (``Who is the strongest one in the team?''), the baseline
hallucinates an incorrect character.
The oracle box redirects visual attention to the correct spatial region, but the model
describes the referent only at the level of a perceptual property (``the blue animal''),
failing to resolve the arrow to a named entity.
Only GRPO fine-tuning enables the full inference chain---from arrow location, to object
identity, to the correct named answer (Benny the bull).
In \textit{Example~2} (``What makes you sneeze?''), the hint alone is sufficient: the
arrow unambiguously isolates a single salient object (the pepper shaker), requiring no
entity-level knowledge to name correctly.
Both oracle conditions succeed here, while the no-hint condition fails entirely.

\paragraph{Interpretation.}
These examples reveal a consistent two-stage failure mode in models without GRPO.
First, without the hint, the model does not attend to the pause arrow at all, defaulting
to plausible but incorrect priors from its pre-training distribution.
Second, even when the hint redirects attention, the model can localize a region without
being able to resolve it to a named entity---a gap that GRPO training on pedagogically
structured content bridges by repeatedly supervising the full reasoning chain from
visual evidence to grounded answer.
This suggests that the pause arrow encodes two distinct learning signals: a
\textit{spatial attention} signal (where to look) that is recoverable through prompting,
and an \textit{entity grounding} signal (what it is called) that requires the kind of
structured supervision our dataset provides.
The quantitative results for this ablation are reported in Table~\ref{tab:pause_hint}.

\pausehintquality  

\paragraph{Scaling behavior.}
To test whether scale alone recovers the pause signal, we evaluate the same
conditions on larger untuned models (Table~\ref{tab:pause_hint_large}).
Qwen2.5-VL-72B with prompted hint (60.66\%) trails the GRPO-finetuned 8B
model (67.61\%), confirming that scale partially recovers spatial attention
but does not replace structured supervision for entity grounding.
Qwen2.5-VL-72B does achieve 11.87\% mean IoU for arrow localization vs.\
$<$3\% for the 8B models, indicating that spatial localization benefits more
directly from scale than entity resolution does.

\pausehintlargetable
\localizationtable
\section{Additional Qualitative Results}
\label{sec:additional_qualitative}
Figure~\ref{fig:additional-qualitative} presents two challenging examples from SoSVQA that illustrate distinct failure modes of baseline models. The spatial location example (left) demonstrates partial occlusion: Swiper the fox is only partially visible behind Dora, requiring the model to recognize an object from incomplete visual information. Without GRPO fine-tuning, Qwen2-VL-2B fails to detect the occluded character entirely, while the GRPO-finetuned variant correctly identifies his location consistent with the spatial perception gains reported in Table~\ref{tab:doravqa_main}. The deixis resolution example (right) illustrates a subtler failure mode. The question ``Mickey, what's that?'' cannot be resolved by identifying the most visually salient object in the frame; it requires grounding the referent \textit{what} in the episode's narrative context. Baseline models default to describing visually dominant elements (the colorful bus) or arbitrarily salient objects (a tree stump), answering ``what is visually prominent?'' rather than ``what is Mickey referring to?''. Our GRPO-finetuned models correctly resolve the reference to the drum by grounding the question in the episodic context; the same entity grounding signal that the \textit{context-question-pause-answer} structure supervises directly (Section~\ref{sec:pause_hint}). Together, these examples demonstrate that GRPO fine-tuning on pedagogically structured content improves both spatial perception under occlusion and narrative-grounded reference resolution. \textbf{Additional interactive qualitative results.} We provide an interactive HTML page in the supplementary materials with further visual examples
across all task types. Please open \texttt{supplementary/index.html} in a web browser
to explore the full set of qualitative results beyond the examples shown in this paper.

 
\lstdefinelanguage{PromptPy}{
  basicstyle=\ttfamily\footnotesize,
  keywordstyle=\color{promptkey}\bfseries,
  stringstyle=\color{promptstr},
  commentstyle=\color{promptcom}\itshape,
  morekeywords={role, content, type, image, text, user, system, assistant},
  morestring=[b]',
  morestring=[b]",
  morecomment=[l]{\#},
  showstringspaces=false,
  breaklines=true,
  breakatwhitespace=true,
  columns=fullflexible,
  keepspaces=true,
  literate=
    {<frame_i>}{{\textcolor{promptplaceholder}{\textit{<frame\_i>}}}}{9}
    {<frame_n>}{{\textcolor{promptplaceholder}{\textit{<frame\_n>}}}}{9}
    {<transcript_with_pause_tokens>}{{\textcolor{promptplaceholder}{\textit{<transcript\_with\_pause\_tokens>}}}}{30}
    {<question_text>}{{\textcolor{promptplaceholder}{\textit{<question\_text>}}}}{17}
    {<pause>}{{\textcolor{promptplaceholder}{\textit{<pause>}}}}{7}
}
 
\newtcblisting{promptbox}[2][]{
  listing only,
  listing options={language=PromptPy},
  colback=promptbg,
  colframe=promptframe,
  arc=2pt,
  outer arc=2pt,
  boxrule=0.6pt,
  left=6pt, right=6pt, top=4pt, bottom=4pt,
  title={\small\bfseries #2},
  fonttitle=\bfseries,
  breakable,
  enhanced,
  #1
}

\section{Prompt Templates}
\label{sec:prompts}
 
To support full reproducibility, this section documents the exact message
structures used during GRPO training and multiple-choice evaluation. Visual frames are uniformly
sampled from the video segment associated with each question; placeholders
of the form \textcolor{promptplaceholder}{\textit{<...>}} are filled in
per-example at runtime.
 
\subsection{Frame Sampling}
 
For every training and evaluation example, we uniformly sample $N$ frames
from the video segment $[t_{\text{start}}, t_{\text{end}}]$ associated
with the question, where $t_{\text{start}}$ is the beginning of the
\textit{context} segment and $t_{\text{end}}$ is the end of the
\textit{pause} segment. Sampled frames are denoted
\textcolor{promptplaceholder}{\textit{<frame\_1>}}, \dots,
\textcolor{promptplaceholder}{\textit{<frame\_n>}}, and are passed in
temporal order. We use $N{=}4$ unless otherwise noted; the prompt
structure is independent of $N$.
 
\subsection{GRPO Training Prompt}
 
During GRPO training, the model receives the sampled frames together with
the episode transcript (with \textcolor{promptplaceholder}{\textit{<pause>}}
tokens inserted at scene boundaries to mark the pedagogical pause), the
question, and an open-ended \texttt{Answer:} cue. Free-form generations are
scored against the ground-truth answer $a^{*}$ using the reward defined
in Eq.~\ref{eq:rewardf}.
 
\begin{promptbox}{GRPO Training Message (Open-Ended Generation)}
messages = [
    {
        "role": "user",
        "content": [
            {"type": "image", "image": <frame_1>},
            {"type": "image", "image": <frame_2>},
            # ... uniformly sampled frames in temporal order ...
            {"type": "image", "image": <frame_n>},
            {
                "type": "text",
                "text": (
                    "You are a helpful visual reasoning assistant for kids.\n"
                    "Think step by step, then give a final concise answer.\n"
                    "Context: <transcript_with_pause_tokens>\n"
                    "Question: <question_text>\n"
                    "Answer:"
                ),
            },
        ],
    },
]
\end{promptbox}
 
\subsection{Multiple-Choice Evaluation Prompt}
 
All cross-benchmark evaluations (Video-MME, NExT-QA, MVBench,
MotionBench, TVBench) and the in-domain SoSVQA evaluation use the
multiple-choice format below. The model is constrained to emit a single
digit selecting one of four choices, which we parse directly. As noted in
Section~\ref{sec:grpo}, this deliberate format mismatch with training
isolates transferable reasoning from answer-format memorization.
 
\begin{promptbox}{Multiple-Choice Evaluation Message (Single-Digit Output)}
messages = [
    {
        "role": "user",
        "content": [
            {"type": "image", "image": <frame_1>},
            {"type": "image", "image": <frame_2>},
            # ... uniformly sampled frames in temporal order ...
            {"type": "image", "image": <frame_n>},
            {
                "type": "text",
                "text": (
                    "You are a helpful visual reasoning assistant for kids.\n"
                    "Think step by step and choose one correct choice "
                    "from 0, 1, 2 or 3.\n"
                    "Only return one single digit as the best answer.\n"
                    "Context: <transcript_with_pause_tokens>\n"
                    "Question: <question_text>\n"
                    "Choices: {\"0\": \"choice_A\", \"1\": \"choice_B\", "
                    "\"2\": \"choice_C\", \"3\": \"choice_D\"}"
                ),
            },
        ],
    },
]
\end{promptbox}
 
\subsection{Summary of Prompt Components}
 
\begin{table}[h]
\centering
\small
\setlength{\tabcolsep}{6pt}
\renewcommand{\arraystretch}{1.15}
\begin{tabular}{@{}p{0.22\linewidth} p{0.34\linewidth} p{0.34\linewidth}@{}}
\toprule
\textbf{Component} & \textbf{Training (GRPO)} & \textbf{Evaluation (MCQ)} \\
\midrule
Frames & $N$ uniformly sampled, temporal order & $N$ uniformly sampled, temporal order \\
Context & Transcript with \textit{<pause>} tokens & Transcript with \textit{<pause>} tokens \\
Question & Free-form question text & Free-form question text \\
Choices & \xmark{} (none) & \cmark{} JSON dict, keys \texttt{0}--\texttt{3} \\
Output target & Open-ended answer string & Single digit $\in \{0,1,2,3\}$ \\
Scoring & Reward (Eq.~\ref{eq:rewardf}) vs.\ $a^{*}$ & Exact-match top-1 accuracy \\
\bottomrule
\end{tabular}
\caption{Differences between training and evaluation prompts. The
training prompt is open-ended; the evaluation prompt adds explicit
multiple-choice formatting and constrains the output to a single digit.}
\label{tab:prompt_components}
\end{table}

\section{Data Access and Licensing}
\label{sec:licensing}
SoSVQA does not redistribute any video frames, audio, or raw episode files. All visual
and audio content was collected from publicly available YouTube uploads and remains
hosted by the original rights holders. Users are required to independently obtain access
to the corresponding episodes under YouTube's existing terms of service. We release
episode identifiers, temporal timestamps, question-answer annotations, choices, transcript
span indices, manually annotated bounding box for the pause-segment visual cues along with the code to align these annotations. As the source material consists of publicly broadcast children's educational programming, the datasets do not introduce new privacy risks; nevertheless, SoSVQA is
intended strictly for research and evaluation purposes, and we discourage any use
outside this scope.

\additionalquality

\section{MCQ Generation and Manual Auditing}
\label{sec:mcq_audit}

To convert the open-ended QA pairs extracted from the
\textit{context-question-pause-answer} structure into a multiple-choice
evaluation format, we use Google Gemini~2.5~Flash to (i) simplify the
gold answer into a concise, MCQ-appropriate phrase and (ii) generate three
plausible but incorrect distractors of similar style and length. The text-only
prompt operates on the question, gold answer, and surrounding transcript
context (truncated to the most recent 1{,}000 characters when longer), and
returns a structured JSON object with the simplified answer and distractors.
The exact prompt template is shown below.

\begin{promptbox}{Gemini MCQ Generation Prompt (Text-Only)}
Create a multiple-choice question with <num_distractors> wrong answer options.

VIDEO CONTEXT:
<transcript_context>

QUESTION: <question_text>
ORIGINAL CORRECT ANSWER: <gold_answer>

TASK:
1. First, simplify the original correct answer to be concise and
   MCQ-appropriate (similar length to distractors, clear and direct)
2. Then generate <num_distractors> wrong but plausible answers that match
   the answer style and length

REQUIREMENTS:
- Simplified answer: Short, clear, direct (1-2 sentences max,
  ideally 1 phrase)
- Distractors: Each must be incorrect, believable, simple words for
  ages 3-7, unique, and similar length to the simplified answer

OUTPUT FORMAT - Copy this structure exactly and fill in all fields:
{
    "simplified_answer": "concise version of the correct answer",
    "distractor_1": "your first wrong answer",
    "distractor_2": "your second wrong answer",
    "distractor_3": "your third wrong answer"
}

IMPORTANT:
- Simplify the answer to be concise (like the distractors will be)
- You must provide simplified_answer, distractor_1, distractor_2, AND
  distractor_3. All fields are required.
\end{promptbox}

Generation uses temperature $0.5$, top-$p = 0.95$, and a maximum of $2{,}048$
output tokens. Each candidate MCQ is post-processed to enforce
case-insensitive uniqueness between the simplified answer and distractors,
and to discard any item that does not yield at least three unique distractors
after deduplication.

\paragraph{Manual auditing of deictic answers.}
Although Gemini reliably produces fluent distractors, a recurring failure mode
arises from the pedagogical structure itself: many answers in the source
material are \textit{deictic}---phrases such as ``there it is,'' ``here it
is,'' or ``right there'' that derive their meaning from the on-screen visual
hint (e.g.,\ a directional arrow or character gesture) rather than from any
named entity. When such a phrase becomes the gold MCQ option alongside three
named-entity distractors generated from the transcript, the resulting item is
trivially solvable by surface form alone: the deictic phrase is the only
choice that does not name a specific entity, leaking the correct answer
without requiring any visual reasoning.

To eliminate this shortcut, we manually audited every MCQ whose simplified
answer or original answer contained a deictic marker (\textit{there it is},
\textit{here it is}, \textit{right there}, and minor morphological variants).
The audit identified \textbf{106 affected items in DoraVQA and 31 in
ClubHVQA}. For each affected item, we manually rewrote the four choices so
that all options share a consistent surface form---typically a yes/no
acknowledgement paired with a directional or perceptual claim---forcing the
model to ground the answer in the visual scene rather than in lexical type.
Figure~\ref{fig:mcq-audit} shows a representative before/after example.

\mcqauditfigure

Concretely, the question \textit{``Now, where's the quiet forest?\ Do you see
the quiet forest?''} originally paired the deictic gold answer
\textit{``There it is!''} against three named-location distractors
(\textit{Little bird?}, \textit{Yellow Valley?}, \textit{Big mountain?}), so
the answer could be selected purely by recognizing which option is not a
named location. After auditing, the four choices become parallel
yes/no--directional statements (\textit{Yes, it's on the left.}, \textit{No,
I can't see it.}, \textit{Yes, it's on the right.}, \textit{I see flowers.}),
which can only be resolved by attending to the on-screen arrow during the
pause segment.

This audit affects $106 / 5{,}348 = 1.98\%$ of DoraVQA and
$31 / 4{,}652 = 0.67\%$ of ClubHVQA. The lower rate in ClubHVQA is consistent
with its more visually-driven, object-selection-dominated structure
(78\% object selection; Figure~\ref{fig:sosvqa-overview}), in which gold
answers are typically named objects rather than deictic references.
All audited items are released as part of the SoSVQA annotations, with both
the original and revised choice sets preserved for transparency.

\section{Limitations}
\label{sec:limit}
While SoSVQA demonstrates that pedagogical structure enables stronger spatial reasoning abilities for VLMs despite limited data scale, the approach has several limitations. First, the dataset depends on the availability and quality of publicly hosted videos and corresponding transcripts; oftentimes the alignment between SRT files and episode timestamps is imperfect and requires human input. Misaligned or noisy subtitles can introduce extraction errors, especially in fast-paced scenes or episodes with inconsistent caption timing. Second, SoSVQA currently draws from only two children’s educational programs--Dora the Explorer and Mickey Mouse Clubhouse. Although these shows share a strong context-question-pause-answer structure, they represent a narrow slice of pedagogical video, and broader generalization would require incorporating additional sources with diverse reasoning styles and visual environments. Third, despite improvements from GRPO fine-tuning, counting performance remains a persistent challenge: models still struggle with fine‑grained enumeration, and the gains observed in counting tasks are modest relative to other reasoning categories. Fourth, due to limited hardware availability, we evaluate only a small set of open-source VLM families and parameter scales. Larger models and alternative architectures may benefit more from structured supervision, but we were unable to explore these directions. Together, these limitations highlight opportunities for expanding the dataset, improving alignment quality, and scaling evaluations to a wider range of model families and pedagogical sources.

\section{Compute Resource}
\label{sec:compute}
For the Qwen2-7B and Qwen3-8B models, all GRPO fine-tuning runs were trained for 450 epochs on a single NVIDIA H200 GPU with an 8-core CPU and 64 GB of system memory. Each model used a combined batch size of 8, gradient accumulation of 4, and 8 sampled generations per prompt. Optimization followed standard GRPO settings, with a learning rate of 1e-4, KL-regularization coefficient $\beta$=0.01, and reward weight of 2.0, informed by our parameter sweep results. For the 2B variants, training was performed on a single NVIDIA A100 GPU, while all other hyperparameters and training configurations remained unchanged. Each individual fine-tuning run required approximately 7-8 GPU hours on the H200 or A100, depending on the model size and checkpoint‑resume behavior. Across all models, ablations, and joint-training configurations, the total compute used for the experiments reported in the paper amounts to roughly 120-150 GPU-hours. The full research project required additional compute beyond the final reported experiments, including preliminary GRPO reward weight sweeps, early alignment-quality diagnostics, and several failed or discarded runs; these exploratory stages added an estimated 40-60 GPU-hours.

Inference was conducted using the default settings provided by LMMS-Eval \cite{lmms_eval2024} and VLLM \cite{kwon2023efficient}, running across a heterogeneous pool of GPUs--including NVIDIA V100, A100, A5000, A6000, H200, and L40S--selected based on model size and input sequence length. Storage requirements were modest: each run required approximately 50-80 GB of temporary workspace for checkpoints, logs, and intermediate outputs. All experiments were conducted on an private high-performance computing (HPC) cluster equipped with GPU nodes accessible through a shared scheduling system.

\section{Safeguards}
\label{sec:sage}
In our work, we take several safeguards to ensure responsible use and to minimize risks of misuse or unintended deployment. First, we do not redistribute any copyrighted video frames, audio, or raw episode content. All visual and audio material remains hosted by the original rights holders on YouTube, and users must independently obtain access under YouTube’s existing terms of service. We release only episode identifiers, timestamps, transcript spans, question–answer annotations, and manually annotated bounding boxes, which pose minimal risk and contain no personally identifiable information. Second, because the source material consists of publicly broadcast children’s educational programming, the dataset does not introduce new privacy risks; nonetheless, we explicitly state that SoSVQA is intended strictly for research and evaluation, and we discourage any use outside this scope.

To further reduce misuse risk, we will provide clear documentation describing the dataset’s structure, intended purpose, and limitations, and we avoid releasing any large-scale pre-trained or instruction-tuned models that could be repurposed for harmful applications. Our fine-tuned models are trained only on pedagogically structured QA pairs and do not include any safety-critical capabilities. Finally, we will include usage guidelines in the repository emphasizing that the dataset should not be used for commercial deployment, content generation, or applications involving children, and we provide scripts that require users to supply their own video access, preventing accidental redistribution of copyrighted or sensitive material.

Together, these safeguards ensure that the released assets are responsibly scoped, legally compliant, and suitable for research use without enabling high-risk downstream applications.

\newpage
\end{document}